%% file: root.tex
\documentclass[10pt,twocolumn,letterpaper]{article}

\newif\ifarxiv
\newif\ifcvpr
\newif\ifcvprfinal

\cvprfalse \cvprfinalfalse \arxivtrue %

\ifcvprfinal
\usepackage{cvpr} %
\fi
\ifcvpr
\usepackage[review]{cvpr} %
\fi
\ifarxiv
\usepackage[pagenumbers]{cvpr} %
\fi

\input{includes/packages}

\input{includes/colors_and_icons}

\input{includes/macros}

\newcommand{\coolname}{\textit{SplaTAM}}
\newcommand{\coolnamefast}{\textit{SplaTAM-S}}

\newcommand{\webpage}{https://spla-tam.github.io}
\newcommand{\webtext}{spla-tam\hspace{0.1em}.\hspace{0.1em}github\hspace{0.1em}.\hspace{0.1em}io}
\newcommand{\codelink}{https://github.com/spla-tam/SplaTAM}

\newcommand{\authorhref}[3][citepurple]{\href{#2}{\color{#1}{#3}}}

\def\paperID{7620} %
\def\confName{CVPR}
\def\confYear{2024}

\ifcvprfinal
\title{\coolname: Splat, Track \& Map 3D Gaussians for Dense RGB-D SLAM \\[1pt]
\Large{\href{\webpage}{\color{Green!90}{\webtext}}}
\vspace{-1.4em}
}
\fi
\ifcvpr
\title{\coolname: Splat, Track \& Map 3D Gaussians for Dense RGB-D SLAM}
\fi
\ifarxiv
\title{\coolname: Splat, Track \& Map 3D Gaussians for Dense RGB-D SLAM \\[1pt]
\Large{\href{\webpage}{\color{Green!90}{\webtext}}}
\vspace{-1.4em}
}
\fi

\author{
\authorhref{https://nik-v9.github.io/}{Nikhil Keetha}$^{1}$,
\authorhref{https://jaykarhade.github.io/}{Jay Karhade}$^{1}$, 
\authorhref{https://krrish94.github.io/}{Krishna Murthy Jatavallabhula}$^{2}$,
\authorhref{https://gengshan-y.github.io/}{Gengshan Yang}$^{1}$, \\
\authorhref{https://theairlab.org/team/sebastian/}{Sebastian Scherer}$^{1}$,
\authorhref{https://www.cs.cmu.edu/~deva/}{Deva Ramanan}$^{1}$, and
\authorhref{https://www.vision.rwth-aachen.de/person/216/}{Jonathon Luiten}$^{1}$
\\[3pt]
$^{1}$\href{https://www.ri.cmu.edu/}{CMU} \hspace{0.5em}
$^{2}$\href{https://www.csail.mit.edu/}{MIT}
}

\begin{document}

\twocolumn[{
\maketitle
\vspace{-1.2em}
\begin{center}
    \centering
    \vspace{-0.3in}
    \includegraphics[trim={0cm 7cm 0.5cm 0cm},clip,width=0.95\linewidth]{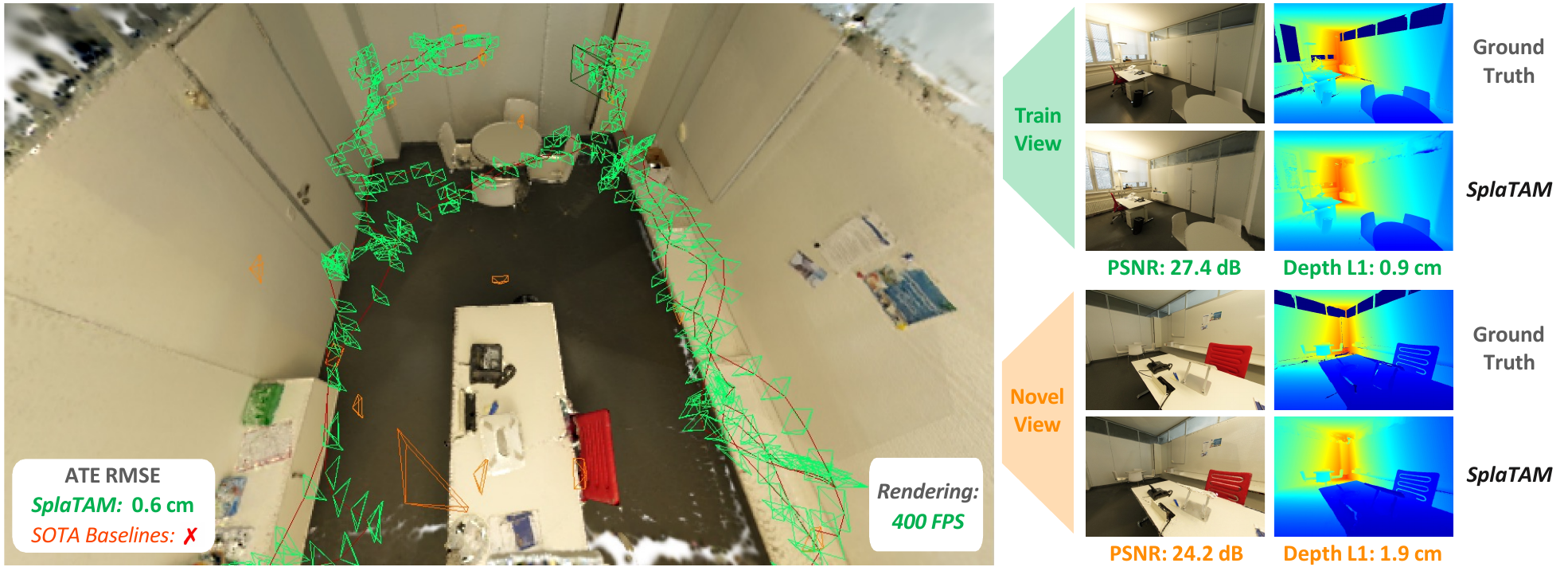}
    \captionof{figure}{\textbf{\coolname} enables \emph{precise camera tracking} and \emph{high-fidelity reconstruction} for dense simultaneous localization and mapping (SLAM) in challenging real-world scenarios. \coolname{} achieves this by online optimization of an explicit volumetric representation, $3$D Gaussian Splatting~\cite{3dgs}, using differentiable rendering. \textit{Left:} We showcase the high-fidelity $3$D Gaussian Map along with the train (SLAM-input) \& novel view camera frustums. It can be noticed that \coolname{} achieves \textit{sub-cm} localization despite the large motion between subsequent cameras in the texture-less environment. This is particularly challenging for state-of-the-art baselines leading to the failure of tracking. \textit{Right:} \coolname{} enables \emph{photo-realistic} rendering of both train \& novel views at \emph{400 FPS} for a resolution of $876 \times 584$.}
    \label{fig:teaser}
\end{center}
}]

\input{text/00_abstract}
\input{text/01_intro}

\input{text/02_related}

\input{text/03_method}

\input{text/04_experiments}

\input{text/05_conclusion}

\vspace{1em}

{   
    \footnotesize
    \noindent \textbf{Acknowledgments.} This work was supported in part by the Intelligence Advanced Research Projects Activity (IARPA) via Department of Interior/ Interior Business Center (DOI/IBC) contract number 140D0423C0074. The U.S. Government is authorized to reproduce and distribute reprints for Governmental purposes notwithstanding any copyright annotation thereon. Disclaimer: The views and conclusions contained herein are those of the authors and should not be interpreted as necessarily representing the official policies or endorsements, either expressed or implied, of IARPA, DOI/IBC, or the U.S. Government. 
    This work used Bridges-2 at PSC through allocation cis220039p from the Advanced Cyberinfrastructure Coordination Ecosystem: Services \& Support (ACCESS) program which is supported by NSF grants \#2138259, \#2138286, \#2138307, \#2137603, and \#213296, and also supported by a hardware grant from Nvidia.
    
    The authors thank Yao He for his support with testing ORB-SLAM3 \& Jeff Tan for testing the code \& demos.
    We also thank Chonghyuk (Andrew) Song, Margaret Hansen, John Kim \& Michael Kaess for their insightful feedback \& discussion on initial versions of the work.
    \par
}

\ifarxiv
\input{supplementary/0_content}
\fi

{
    \small
    \bibliographystyle{ieeenat_fullname}
    \bibliography{root}
}

\ifcvprfinal
\input{supplementary/0_content}
\fi

\end{document}

%% file: includes/packages.tex
\usepackage{booktabs}
\usepackage{graphics} %
\usepackage{graphicx}
\usepackage{pifont} %
\usepackage{epsfig} %
\usepackage{mathptmx} %
\usepackage{times} %
\usepackage{amsmath} %
\usepackage{amssymb}  %
\usepackage{arydshln} %
\usepackage{capt-of}
\usepackage{textcomp}  %
\usepackage[dvipsnames]{xcolor}
\usepackage{xcolor,colortbl}
\usepackage[accsupp]{axessibility}  %
\usepackage{cite}
\usepackage{multirow, makecell}
\usepackage{array}
\usepackage{bm}
\definecolor{citepurple}{rgb}{0.288,0.1196,0.7}
\usepackage[pagebackref,breaklinks,colorlinks,bookmarks,citecolor=citepurple]{hyperref}

%% file: includes/colors_and_icons.tex
\definecolor{Gray}{gray}{0.50}
\newcolumntype{g}{>{\columncolor{Gray}}c}
\definecolor{ffe1da}{RGB}{255,225,218}
\definecolor{F7E0D5}{RGB}{247,224,213}
\definecolor{darkF7E0D5}{RGB}{209,154,128}
\colorlet{Light}{White!0!F7E0D5}

\colorlet{tabfirst}{Green!25}
\definecolor{tabthird}{rgb}{1, 0.85, 0.7}
\definecolor{tabsecond}{rgb}{1, 0.96, 0.7}

\newcommand{\greencheck}{{\color{green}\checkmark}}
\newcommand{\redx}{{\color{red}\ding{55}}}

%% file: includes/macros.tex
\definecolor{darkorange}{rgb}{1.0, 0.54, 0}
\definecolor{darkpurple}{rgb}{0.288,0.1196,0.7}

\definecolor{amber}{rgb}{1.0, 0.75, 0.0}

\newcommand{\Rom}[1]{\uppercase\expandafter{\romannumeral #1\relax}}

\newcommand{\higher}{$\uparrow$}
\newcommand{\perflower}{$\downarrow$}

\definecolor{darkgray}{rgb}{0.2, 0.2, 0.2}

\usepackage[capitalize]{cleveref}
\crefname{section}{Section}{Sections}
\crefname{table}{Table}{Tables}

\makeatletter
\newcommand{\cdashmidrule}[1]{%
  \noalign{\vskip\aboverulesep}
  \cdashline{#1}
  \noalign{\vskip\belowrulesep}}
\makeatother

\newcommand*{\subfigref}[2][]{%
  Fig. \hyperref[{fig:#2}]{%
    \ref*{fig:#2}%
    \ifx\\#1\\%
    \else
      \,#1%
    \fi
  }%
}

\makeatletter

\newcommand*{\@rowstyle}{}

\newcommand*{\rowstyle}[1]{%
  \gdef\@rowstyle{#1}%
  \@rowstyle\ignorespaces%
}

\newcolumntype{=}{%
  >{\gdef\@rowstyle{}}%
}

\newcolumntype{+}{%
  >{\@rowstyle}%
}

\makeatother

%% file: text/00_abstract.tex
\begin{abstract}
\vspace{-1.2em}

Dense simultaneous localization and mapping (SLAM) is crucial for robotics and augmented reality applications. However, current methods are often hampered by the non-volumetric or implicit way they represent a scene. This work introduces \coolname{}, an approach that, for the first time, leverages explicit volumetric representations, i.e., 3D Gaussians, to enable high-fidelity reconstruction from a single unposed RGB-D camera, surpassing the capabilities of existing methods. \coolname{} employs a simple online tracking and mapping system tailored to the underlying Gaussian representation. It utilizes a silhouette mask to elegantly capture the presence of scene density. This combination enables several benefits over prior representations, including fast rendering and dense optimization, quickly determining if areas have been previously mapped, and structured map expansion by adding more Gaussians. Extensive experiments show that \coolname{} achieves up to $2 \times$ superior performance in camera pose estimation, map construction, and novel-view synthesis over existing methods, paving the way for more immersive high-fidelity SLAM applications.

\end{abstract}

%% file: text/01_intro.tex
\newpage

\section{Introduction}
\label{sec:intro}

Visual simultaneous localization and mapping (SLAM)---the task of estimating the pose of a vision sensor (such as a depth camera) \textit{and} a map of the environment---is an essential capability for vision or robotic systems to operate in previously unseen 3D environments.
For the past three decades, SLAM research has extensively centered around the question of \emph{map representation} -- resulting in a variety of sparse~\cite{slam-past-present-future, orbslam2, orbslam3, monoslam}, dense~\cite{lsdslam, curless1996volumetric, steinbrucker2011real, kinectfusion, dtam, pointfusion, kerl2013vo, kintinuous, elasticfusion, bundlefusion}, and neural scene representations~\cite{nerf, imap, voxfusion, zhu2023nicer, sandstrom2023point, rosinol2022nerf}.
Map representation is a fundamental choice that dramatically impacts the design of every processing block within the SLAM system, as well as of the downstream tasks that depend on the outputs of SLAM.

In terms of dense visual SLAM, the most successful \emph{explicit} \& \emph{handcrafted} representations are points, surfels/flats, and signed distance fields.
While systems based on such map representations have matured to production level over the past years, there are still significant shortcomings that need to be addressed.
Tracking explicit representations relies crucially on the availability of rich 3D geometric features and high-framerate captures.
Furthermore, these approaches only reliably explain the observed parts of the scene with dense view coverage; many applications, such as mixed reality and high-fidelity 3D capture, require techniques that can also explain/synthesize unobserved/novel camera viewpoints~\cite{sarlin2022lamar}.

The shortcomings of \emph{explicit} representations, combined with the emergence of radiance field-based volumetric representations~\cite{nerf} for high-quality image synthesis, have fueled a recent class of methods that attempt to encode the scene into the weight space of a neural network.
These \emph{implicit} \& \emph{volumetric} SLAM algorithms~\cite{nice-slam, sandstrom2023point} benefit from high-fidelity global maps and image reconstruction losses that capture dense photometric information via differentiable rendering.
However, these methods have several issues in the SLAM setting - they are computationally inefficient, not easy to edit, do not model spatial geometry explicitly, and suffer from catastrophic forgetting.

In this context, we explore the question, ``\textbf{How can an explicit volumetric representation benefit the design of a SLAM solution?}"
Specifically, we use an explicit volumetric representation based on 3D Gaussians~\cite{3dgs} to \emph{Splat (Render)}, \emph{Track}, and \emph{Map} for SLAM.
We find that this leads to the following benefits over existing map representations:
\begin{itemize}
    \item \textbf{Fast rendering and dense optimization:} 3D Gaussians can be rendered as images at speeds up to $400$ FPS, making them significantly faster to optimize than the implicit \& volumetric alternatives. The key enabling factor for this fast optimization is the rasterization of 3D primitives. We introduce several simple modifications that make splatting even faster for SLAM, including the removal of view-dependent appearance and the use of isotropic (spherical) Gaussians. Furthermore, this allows us to use dense photometric loss for SLAM in real-time, in contrast to prior explicit \& implicit map representations that rely respectively on sparse 3D geometric features or pixel sampling to maintain efficiency.
    \item \textbf{Maps with explicit spatial extent:} The existing parts of a scene, i.e., the spatial frontier, can be easily identified by rendering a silhouette mask, which represents the accumulated opacity of the 3D Gaussian map for a particular view. Given a new view, this allows one to efficiently identify which portions of the scene are new content (outside the map's spatial frontier). This is crucial for camera tracking as we only want to use mapped parts of the scene for the optimization of the camera pose. Furthermore, this enables easy map updates, where the map can be expanded by adding Gaussians to the unseen regions while still allowing high-fidelity rendering. In contrast, for prior implicit \& volumetric map representations, it is not easy to update the map due to their interpolatory behavior, while for explicit non-volumetric representations, the photo-realistic \& geometric fidelity is limited.
    \item \textbf{Direct optimization of scene parameters:} As the scene is represented by Gaussians with physical 3D locations, colors, and sizes, there is a direct, almost linear (projective) gradient flow between the parameters and the dense photometric loss.
    Because camera motion can be thought of as keeping the camera still and moving the scene, we also have a direct gradient into the camera parameters, which enables fast optimization. Prior differentiable (implicit \& volumetric) representations don't have this, as the gradient needs to flow through (potentially many) non-linear neural network layers.
\end{itemize}

\vspace{0.4em} %

\noindent Given all of the above advantages, an explicit volumetric representation is a natural way for efficiently inferring a high-fidelity spatial map while simultaneously estimating camera motion, as is also visible from \cref{fig:teaser}.
We show across all our experiments on simulated and real data that our approach, \coolname{}, achieves state-of-the-art results compared to all previous approaches for camera pose estimation, map reconstruction, and novel-view synthesis.

%% file: text/02_related.tex
\input{figs/approach}

\section{Related Work}
\label{sec:related}

In this section, we briefly review various approaches to dense SLAM, with a particular emphasis on recent approaches that leverage implicit representations encoded in overfit neural networks for tracking and mapping.
For a more detailed review of traditional SLAM methods, we refer the interested reader to this excellent survey~\cite{slam-past-present-future}.

\vspace{0.5em} %

\noindent \textbf{Traditional approaches to dense SLAM} have explored a variety of explicit representations, including 2.5D images~\cite{steinbrucker2011real,kerl2013vo}, (truncated) signed-distance functions~\cite{kinectfusion,kintinuous,dtam,bundlefusion}, gaussian mixture models \cite{goel2023incremental, goel_probabilistic_2023} and circular surfels~\cite{pointfusion,elasticfusion,bad-slam}.
Of particular relevance to this work is \emph{surfels}, which are colored circular surface elements and can be optimized in real-time from RGB-D image inputs as shown in~\cite{pointfusion, elasticfusion, bad-slam}.
While the aforementioned SLAM approaches do not assume the visibility function to be differentiable, there exist modern differentiable rasterizers that enable gradient flow through depth discontinuities~\cite{diff-surface-splatting}.
However, since 2D \emph{surfels} are discontinuous, they need careful regularization to prevent holes~\cite{bad-slam, diff-surface-splatting}.
In this work, we use volumetric (as opposed to surface-only) scene representations in the form of 3D Gaussians, enabling easy continuous optimization for fast and accurate SLAM.

\vspace{0.5em} %

\noindent \textbf{Pretrained neural network representations} have been integrated with traditional SLAM techniques, largely focusing on predicting depth from RGB images. These approaches range from directly integrating depth predictions from a neural network into SLAM~\cite{cnnslam}, to learning a variational autoencoder that decodes compact optimizable latent codes to depth maps~\cite{codeslam,scenecode}, to approaches that simultaneously learn to predict a depth cost volume and tracking~\cite{deeptam}.

\vspace{0.2em} %

\noindent \textbf{Implicit scene representations}: 
iMAP~\cite{imap} first performed both tracking and mapping using neural implicit representations. To improve scalability, NICE-SLAM~\cite{nice-slam} proposed the use of hierarchical multi-feature grids. On similar lines, iSDF~\cite{ortiz2022isdf} used implicit representations to efficiently compute signed-distances, compared to~\cite{kinectfusion, oleynikova2017voxblox}. Following this, a number of works~\cite{zhu2023nicer, wang2023co, li2023dense, ming2022idf, sandstrom2023uncle, zhang2023hi, han2023ro, kong2023vmap, rosinol2022nerf}, have recently advanced implicit-based SLAM through a number of ways - by reducing catastrophic forgetting via continual learning (experience replay), capturing semantics, incorporating uncertainty, employing efficient resolution hash-grids and encodings, and using improved losses. More recently, Point-SLAM~\cite{sandstrom2023point}, proposes an alternative route, similar to \cite{xu2022point} by using neural point clouds and performing volumetric rendering with feature interpolation, offering better 3-D reconstruction, especially for robotics. However, like other implicit representations, volumetric ray sampling greatly limits its efficiency, thereby resorting to optimization over a sparse set of pixels as opposed to per-pixel dense photometric error. Instead, \coolname{}'s explicit volumetric radiance model leverages fast rasterization, enabling complete use of per-pixel dense photometric errors.

\noindent \textbf{3D Gaussian Splatting}:
Recently, 3D Gaussians have emerged as a promising 3D scene representation~\cite{3dgs, keselman2022approximate, wang2022voge, keselman2023flexible}, in particular with the ability to differentiably render them extremely quickly via splatting~\cite{3dgs}. This approach has also been extended to model dynamic scenes~\cite{luiten2023dynamic, wu20234d, yang2023deformable, yang2023real} with dense 6-DOF motion~\cite{luiten2023dynamic}.
Such approaches for both static and dynamic scenes require that each input frame has an accurately known 6-DOF camera pose, in order to successfully optimize the representation. For the first time our approach removes this constraint, simultaneously estimating the camera poses while also fitting the underlying Gaussian representation.

%% file: figs/approach.tex
\begin{figure*}
    \centering
    \includegraphics[trim={0cm 4cm 0cm 0cm},clip,width=0.95\linewidth]{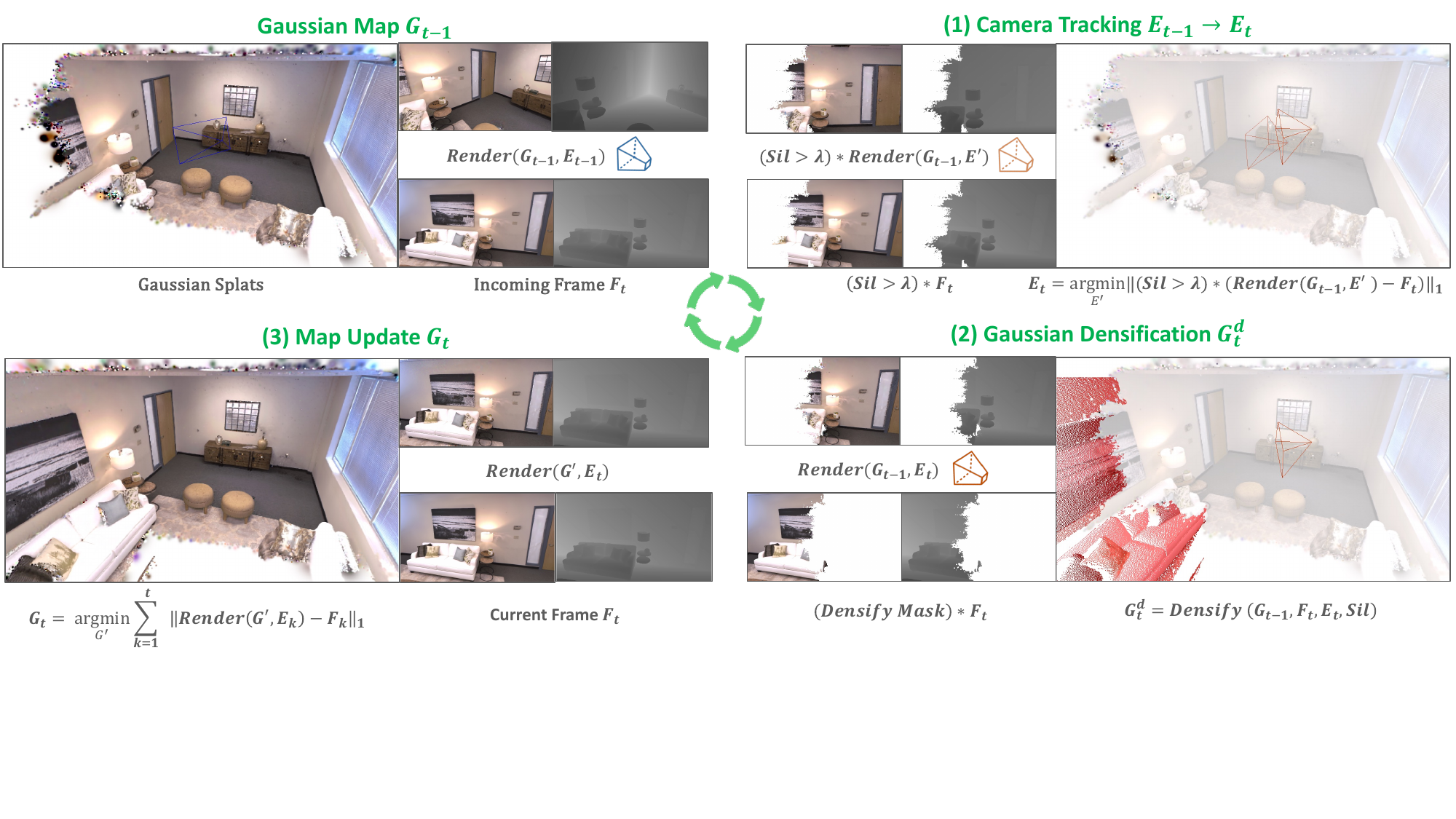}
    \caption{\textbf{Overview of \coolname{}.} \textit{Top-Left:} The input to our approach at each timestep is the current RGB-D frame and the 3D Gaussian Map Representation that we have built representing all of the scene seen so far. \textit{Top-Right:} Step (1) estimates the camera pose for the new image, using silhouette-guided differentiable rendering. \textit{Bottom-Right:} Step (2) increases the map capacity by adding new Gaussians based on the rendered silhouette and input depth. \textit{Bottom-Left:} Step (3) updates the underlying Gaussian map with differentiable rendering.}
    \label{fig:pipeline}
\end{figure*}

%% file: text/03_method.tex
\section{Method}
\label{sec:approach}

\coolname{} is the first dense RGB-D SLAM solution to use $3$D Gaussian Splatting~\cite{3dgs}. By modeling the world as a collection of 3D Gaussians which can be rendered into high-fidelity color and depth images, we are able to directly use differentiable-rendering and gradient-based-optimization to optimize both the camera pose for each frame and an underlying volumetric discretized map of the world.

\paragraph{Gaussian Map Representation.}
We represent the underlying map of the scene as a set of 3D Gaussians. We made a number of simplifications to the representation proposed in~\cite{3dgs}, by using only view-independent color and forcing Gaussians to be isotropic. This implies each Gaussian is parameterized by only $8$ values: three for its RGB color ${\bf c}$, three for its center position $\bm{\mu} \in \mathbb{R}^3$, one for its radius $r$ and one for its opacity $o \in [0,1]$. %
Each Gaussian influences a point in 3D space $\mathbf{x} \in \mathbb{R}^3$ %
according to the standard (unnormalized) Gaussian equation weighted by its opacity:
\begin{align}
    f(\mathbf{x}) = o \exp\left(-\frac{\|\mathbf{x} - \boldsymbol{\mu}\|^2}{2r^2}\right).\label{eq:gauss}
\end{align}
\paragraph{Differentiable Rendering via Splatting.}
The core of our approach is the ability to render high-fidelity color, depth, and silhouette images from our underlying Gaussian Map into any possible camera reference frame in a differentiable way. This differentiable rendering allows us to directly calculate the gradients in the underlying scene representation (Gaussians) and camera parameters with respect to the error between the renders and provided RGB-D frames, and update both the Gaussians and camera parameters to minimize this error, thus fitting both accurate camera poses and an accurate volumetric representation of the world. 

Gaussian Splatting~\cite{3dgs} renders an RGB image as follows: Given a collection of 3D Gaussians and camera pose, first sort all Gaussians from front-to-back. RGB images can then be efficiently rendered by alpha-compositing the splatted 2D projection of each Gaussian in order in pixel space. The rendered color of pixel ${\bf p} = (u,v)$ can be written as:
\begin{equation}
C(\mathbf{p}) = \sum_{i=1}^{n} \mathbf{c}_i f_i(\mathbf{p}) \prod_{j=1}^{i-1} (1 - f_j(\mathbf{p})),
\end{equation}
where $f_i(\bf{p})$ %
is computed as in Eq.~\eqref{eq:gauss} but with $\boldsymbol{\mu}$ and $r$ of the splatted 2D Gaussians in pixel-space: %
\begin{equation}
    \boldsymbol{\mu}^{\textrm{2D}} = K \frac{E_t \boldsymbol{\mu}}{d}, \;\;\;\;\;\;\;\;\;\;
r^{\textrm{2D}} = \frac{f r}{d}, %
\quad \text{where} \quad d = (E_t \boldsymbol{\mu})_z.
\end{equation}
Here, $K$ is the (known) camera intrinsic matrix, $E_t$ is the extrinsic matrix capturing the rotation and translation of the camera at frame $t$, $f$ is the (known) focal length, and $d$ is the depth of the $i^{th}$ Gaussian in camera coordinates.%

We propose to similarly differentiably render depth:

\begin{equation}
    D(\mathbf{p}) = \sum_{i=1}^{n} d_i f_i(\mathbf{p}) \prod_{j=1}^{i-1} (1 - f_j(\mathbf{p})),
\end{equation}
which can be compared against the input depth map and return gradients with respect to the 3D map. 

We also render a silhouette image to determine visibility -- e.g. if a pixel contains information from the current map:

\begin{equation}
    S(\mathbf{p}) = \sum_{i=1}^{n} f_i(\mathbf{p}) \prod_{j=1}^{i-1} (1 - f_j(\mathbf{p})).
\end{equation}

\paragraph{SLAM System.}
We build a SLAM system from our Gaussian representation and differentiable renderer. We begin with a brief overview and then describe each module in detail. Assume we have an existing map (represented via a set of 3D Gaussians) that has been fitted from a set of camera frames 1 to t. Given a new RGB-D frame $t+1$, our SLAM system performs the following steps (see \cref{fig:pipeline}):
\begin{enumerate}[(1)]
    \item {\bf Camera Tracking}. We minimize the image and depth reconstruction error of the RGB-D frame with respect to camera pose parameters for $t+1$, but only evaluate errors over pixels within the visible silhouette. %
    \item {\bf Gaussian Densification.} We add new Gaussians to the map based on the rendered silhouette and input depth.
    \item {\bf Map Update.} Given the camera poses from frame 1 to $t+1$, we update the parameters of all the Gaussians in the scene by minimizing the RGB and depth errors over all images up to $t+1$. In practice, to keep the batch size manageable, a selected subset of keyframes that overlap with the most recent frame are optimized.
\end{enumerate}

\paragraph{Initialization.} For the first frame, the tracking step is skipped, and the camera pose is set to identity. In the densification step, since the rendered silhouette is empty, all pixels are used to initialize new Gaussians. Specifically, for each pixel, we add a new Gaussian with the color of that pixel, the center at the location of the unprojection of that pixel depth, an opacity of $0.5$, and a radius equal to having a one-pixel radius upon projection into the 2D image given by dividing the depth by the focal length:
\begin{align}
r = \frac{D_{\textrm{GT}}}{f}
\end{align}

\paragraph{Camera Tracking.}
Camera Tracking aims to estimate the camera pose of the current incoming online RGB-D image. 
The camera pose is initialized for a new timestep by a constant velocity forward projection of the pose parameters in the camera center + quaternion space. E.g. the camera parameters are initialized using the following:
\begin{align}
E_{t+1} = E_t + (E_t - E_{t \text{-} 1})
\end{align}
The camera pose is then updated iteratively by gradient-based optimization through differentiably rendering RGB, depth, and silhouette maps, and updating the camera parameters to minimize the following loss while keeping the Gaussian parameters fixed:
\begin{align}
L_{\textrm{t}} = \sum_{\textbf{p}} \Bigl(S(\textbf{p}) > 0.99\Bigr) \Bigl(\textrm{L}_1\bigl(D(\textbf{p})\bigr) + 0.5 \textrm{L}_1\bigl(C(\textbf{p})\bigr)\Bigr)
\end{align}
This is an L1 loss on both the depth and color renders, with color weighted down by half.
The color weighting is empirically selected, where we observe the range of $C(\textbf{p})$ to be [0.01, 0.03] \& $D(\textbf{p})$ to be [0.002, 0.006].
We only apply the loss over pixels that are rendered from well-optimized parts of the map by using our rendered visibility \emph{silhouette} which captures the epistemic uncertainty of the map.
This is very important for tracking new camera poses, as often new frames contain new information that hasn't been captured or well optimized in our map yet. 
The L1 loss also gives a value of 0 if there is no ground-truth depth for a pixel.

\paragraph{Gaussian Densification.}
Gaussian Densification aims to initialize new Gaussians in our Map at each incoming online frame. After tracking we have an accurate estimate for the camera pose for this frame, and with a depth image we have a good estimate for where Gaussians in the scene should be. However, we don't want to add Gaussians where the current Gaussians already accurately represent the scene geometry. Thus we create a densification mask to determine which pixels should be densified:
\begin{multline}
M(\textbf{p}) = \Bigl(S(\textbf{p}) < 0.5\Bigr) + \\ 
\Bigl( D_{\textrm{GT}}(\textbf{p}) < D(\textbf{p}) \Bigr) \Bigl(\textrm{L}_1\bigl(D(\textbf{p})\bigr) > \lambda \textrm{MDE} \Bigr)
\end{multline}
This mask indicates where the map isn't adequately dense ($S < 0.5$), or where there should be new geometry in front of the current estimated geometry (i.e., the ground-truth depth is in front of the predicted depth, and the depth error is greater than $\lambda$ times the median depth error (MDE), where $\lambda$ is empirically selected as $50$).
For each pixel, based on this mask, we add a new Gaussian following the same procedure as first-frame initialization.%

\paragraph{Gaussian Map Updating.}
This aims to update the parameters of the 3D Gaussian Map given the set of online camera poses estimated so far. This is done again by differentiable-rendering and gradient-based-optimization, however unlike tracking, in this setting the camera poses are fixed, and the parameters of the Gaussians are updated.

This is equivalent to the ``classic" problem of fitting a radiance field to images with known poses. However, we make two important modifications. Instead of starting from scratch, we warm-start the optimization from the most recently constructed map. We also do not optimize over all previous (key)frames but select frames that are likely to influence the newly added Gaussians.
We save each $n^{\text{th}}$ frame as a keyframe and select $k$ frames to optimize, including the current frame, the most recent keyframe, and $k-2$ previous keyframes which have the highest overlap with the current frame. Overlap is determined by taking the point cloud of the current frame depth map and determining the number of points inside the frustum of each keyframe. %

This phase optimizes a similar loss as during tracking, except we don't use the silhouette mask as we want to optimize over all pixels. Furthermore, we add an SSIM loss to RGB rendering and cull useless Gaussians that have near 0 opacity or are too large, as done partly in \cite{3dgs}.

%% file: text/04_experiments.tex
\input{tables/tracking}

\vspace{-0.2em}

\section{Experimental Setup}
\label{sec:setup}

\paragraph{Datasets and Evaluation Settings.}
We evaluate our approach on four datasets: ScanNet++~\cite{scannet++}, Replica~\cite{replica}, TUM-RGBD~\cite{tum-rgbd} and the original ScanNet~\cite{scannet}. The last three are chosen in order to follow the evaluation procedure of previous radiance-field-based SLAM methods Point-SLAM~\cite{sandstrom2023point} and NICE-SLAM~\cite{nice-slam}. However, we also add ScanNet++~\cite{scannet++} evaluation because none of the other three benchmarks have the ability to evaluate rendering quality on hold out \textit{novel} views and only evaluate camera pose estimation and rendering on the training views.

Replica~\cite{replica} is the simplest benchmark as it contains synthetic scenes, highly accurate and complete (synthetic) depth maps, and small displacements between consecutive camera poses. TUM-RGBD~\cite{tum-rgbd} and the original ScanNet are harder, especially for dense methods, due to poor RGB and depth image quality as they both use old low-quality cameras. Depth images are quite sparse with lots of missing information, and the color images have a very high amount of motion blur. For ScanNet++~\cite{scannet++} we use the DSLR captures from two scenes (\texttt{8b5caf3398 (S1)} and \texttt{b20a261fdf (S2)}), where complete dense trajectories are present. In contrast to other benchmarks, ScanNet++ color and depth images are very high quality, and provide a second capture loop for each scene to evaluate completely novel hold-out views. However, each camera pose is very far apart from one another making pose-estimation very difficult. The difference between consecutive frames on ScanNet++ is about the same as a 30-frame gap on Replica. For all benchmarks except ScanNet++, we take baseline numbers from Point-SLAM~\cite{sandstrom2023point}.
Similar to Point-SLAM, we evaluate every $5$th frame for the training view rendering benchmarking on Replica.
Furthermore, for all comparisons to prior baselines, we present results as the average of $3$ seeds (0-2) and use seed 0 for the ablations.

\vspace{-1.2em}

\paragraph{Evaluation Metrics.}
We follow all evaluation metrics for camera pose estimation and rendering performance as~\cite{sandstrom2023point}. For measuring RGB rendering performance we use PSNR, SSIM and LPIPS. For depth rendering performance we use Depth L1 loss. For camera pose estimation tracking we use the average absolute trajectory error (ATE RMSE).

\vspace{-1.2em}

\paragraph{Baselines.}
The main baseline method we compare to is Point-SLAM~\cite{sandstrom2023point}, the previous state-of-the-art (SOTA) method for dense radiance-field-based SLAM.
We also compare to older dense SLAM approaches such as NICE-SLAM~\cite{nice-slam}, Vox-Fusion~\cite{voxfusion}, and ESLAM~\cite{eslam} where appropriate. 
Lastly, we compare against traditional SLAM systems such as Kintinuous~\cite{kintinuous}, ElasticFusion~\cite{elasticfusion}, and ORB-SLAM2~\cite{orbslam2} on TUM-RGBD, DROID-SLAM~\cite{teed2021droid} on Replica, and ORB-SLAM3~\cite{orbslam3} on ScanNet++.

\section{Results \& Discussion}

In this section, we first discuss our evaluation results on camera pose estimation for the four benchmark datasets.
Then, we further showcase our high-fidelity $3$D Gaussian reconstructions and provide qualitative and quantitative comparisons of our rendering quality (both for novel views and input training views).
Finally, we discuss pipeline ablations and provide a runtime comparison.

\vspace{-1.5em}

\paragraph{Camera Pose Estimation Results.}
In Table~\ref{tab:tracking}, we compare our method's camera pose estimation results to a number of baselines on four datasets~\cite{scannet,scannet++,replica,tum-rgbd}. 

On ScanNet++~\cite{scannet++}, both SOTA SLAM approaches Point-SLAM~\cite{sandstrom2023point} and ORB-SLAM3~\cite{orbslam3} (RGB-D variant) completely fail to correctly track the camera pose due to the very large displacement between contiguous cameras, and thus give very large pose-estimation errors.
In particular, for ORB-SLAM3, we observe that the texture-less ScanNet++ scans cause the tracking to re-initialize multiple times due to the lack of features.
In contrast, our approach successfully manages to track the camera over both sequences giving an average trajectory error of only 1.2cm.

On the relatively easy synthetic Replica~\cite{replica} dataset, the de-facto evaluation benchmark, our approach reduces the trajectory error over the prior SOTA-dense baseline~\cite{sandstrom2023point} by more than $30\%$ from $0.52$cm to $0.36$cm.
Furthermore, \coolname{} provides better or competitive performance to a feature-based tracking method such as DROID-SLAM~\cite{teed2021droid}.

On TUM-RGBD~\cite{tum-rgbd}, all the volumetric methods struggle immensely due to both poor depth sensor information (very sparse) and poor RGB image quality (extremely high motion blur). 
Compared to prior methods in this category~\cite{sandstrom2023point, nice-slam}, \coolname{} still significantly outperforms, decreasing the trajectory error of the prior SOTA in this category~\cite{sandstrom2023point} by almost $40\%$, from $8.92$cm to $5.48$cm. 
Also, we observe that feature-based methods (ORB-SLAM2~\cite{orbslam2}) still outperform dense methods on this benchmark.

The original ScanNet~\cite{scannet} benchmark has similar issues to TUM-RGBD, where no dense volumetric method is able to obtain results with less than $10$cm error \& \coolname{} performs similarly to the two prior SOTA methods~\cite{sandstrom2023point, nice-slam}.

Overall these camera pose estimation results are extremely promising and show off the strengths of our \coolname{} method. The results on ScanNet++ show that if you have high-quality clean input images, our approach can successfully and accurately perform SLAM even with extremely large motions between camera positions, which is not something that is possible with previous SOTA approaches such as Point-SLAM~\cite{sandstrom2023point} and ORB-SLAM3~\cite{orbslam3}.

\vspace{-1.2em}

\paragraph{Rendering Quality Results.}

In Table~\ref{tab:replica}, we evaluate our method's rendering quality on the input views of Replica dataset~\cite{replica} (as is the standard evaluation approach from Point-SLAM~\cite{sandstrom2023point} and NICE-SLAM~\cite{nice-slam}). Our approach achieves similar PSNR, SSIM, and LPIPS results as Point-SLAM~\cite{sandstrom2023point}, although the comparison is not fair as Point-SLAM has the unfair advantage as it takes the ground-truth depth of these images as an input for where to sample its 3D volume for rendering. Our approach achieves much better results than the other baselines Vox-Fusion~\cite{voxfusion} and NICE-SLAM~\cite{nice-slam}, improving over both by around $10$dB in PSNR. 

In general, we believe that the rendering results on Replica in Table~\ref{tab:replica} are irrelevant because the rendering performance is evaluated on the same training views that were passed in as input, and methods can simply have a high capacity and overfit to these images.
We only show this as this is the de-facto evaluation for prior methods, and we wish to have some way to compare against them.

\input{tables/replica}

Hence, a better evaluation is to evaluate novel-view rendering. However, all current SLAM benchmarks don't have a hold-out set of images separate from the camera trajectory that the SLAM algorithm estimates, so they cannot be used for this purpose. Therefore, we set up a novel benchmark for this using the new high-quality ScanNet++~\cite{scannet++} dataset.

The results for both novel-view and training-view rendering on this ScanNet++ benchmark can be found in Table~\ref{tab:scannetpp}. Our approach achieves a good novel-view synthesis result of $24.41$ PSNR on average and slightly higher on training views with $27.98$ PSNR. 
Note that for novel-view synthesis, we use the ground-truth pose of the novel view to align it with the origin of the SLAM map, i.e., the first SLAM frame.
Since Point-SLAM~\cite{sandstrom2023point} fails to successfully estimate the camera poses and build a good map, it also completely fails on the task of novel-view synthesis. %

We can also evaluate the geometric reconstruction of the scene by evaluating the rendered depth and comparing it to the ground-truth depth, again for both training and novel views. Our method obtains an incredibly accurate reconstruction with a depth error of only around $2$cm in novel views and $1.3$cm in training views. 

Visual results of both novel-view and training-view rendering for RGB and depth can be seen in Fig.~\ref{fig:rendering}. As can be seen, our methods achieve visually excellent results over both scenes for both novel and training views. In contrast, Point-SLAM~\cite{sandstrom2023point} fails at camera-pose tracking and overfits to the training views, and isn't able to successfully render novel views at all. Point-SLAM also takes the ground-truth depth as an input for rendering to determine where to sample, and as such the depth maps look similar to the ground-truth, while color rendering is completely wrong.

With this novel benchmark that is able to correctly evaluate Novel-View Synthesis and SLAM simultaneously, as well as our approach as a strong initial baseline for this benchmark, we hope to inspire many future approaches that improve upon these results for both tasks.
\input{tables/scannetpp}

\input{tables/color_depth_ablation}
\input{tables/tracking_ablation}

\vspace{-1em}

\input{figs/rendering}

\noindent \paragraph{Color and Depth Loss Ablation.}
Our \coolname{} involves fitting the camera poses (during tracking) and the scene map (during mapping) using both a photometric (RGB) and depth loss. In Table~\ref{tab:color_depth_ablation}, we ablate the decision to use both and investigate the performance of only one or the other for both tracking and mapping. We do this using Room 0 of Replica. With only depth our method completely fails to track the camera trajectory, because the L1 depth loss doesn't provide adequate information in the x-y image plane. Using only an RGB loss successfully tracks the camera trajectory (although with more than 5x the error as using both). Both the RGB and depth work together to achieve excellent results. With only the color loss, reconstruction PSNR is really high, only $1.5$ PSNR lower than the full model. However, the depth L1 is much higher using only color vs directly optimizing this depth error. In the color-only experiments, the depth is not used for tracking \& mapping, but it is used for the densification and initialization of Gaussians. 

\vspace{-1em}

\noindent \paragraph{Camera Tracking Ablation.}
In Table~\ref{tab:tracking_ablation}, we ablate three aspects of our camera tracking: (1) the use of forward velocity propagation, (2) the use of a silhouette mask to mask out invalid map areas in the loss, and (3) setting the silhouette threshold to 0.99 instead of 0.5. 
All three are critical to excellent results. 
Without forward velocity propagation, tracking still works, but the overall error is more than 10x higher, which subsequently degrades the map and rendering. 
Silhouette is critical as without it tracking completely fails. 
Setting the silhouette threshold to 0.99 allows the loss to be applied on well-optimized pixels in the map, thereby leading to an important 5x reduction in the error compared to the threshold of 0.5, which is used for the densification.

\vspace{-1em}

\noindent \paragraph{Runtime Comparison.}
In Table~\ref{tab:runtime}, we compare our runtime to NICE-SLAM~\cite{nice-slam} and Point-SLAM~\cite{sandstrom2023point} on a Nvidia RTX 3080 Ti. Each iteration of our approach renders a full 1200x980 pixel image ($\sim$1.2 mil pixels) to apply the loss for both tracking and mapping. Other methods use only $200$ pixels for tracking and $1000$ pixels for mapping each iteration (but attempt to cleverly sample these pixels). Even though we differentiably render 3 orders of magnitude more pixels, our approach incurs similar runtime, primarily due to the efficiency of rasterizing 3D Gaussians.
Additionally, we show a version of our method with fewer iterations and half-resolution densification, \coolnamefast{}, which works 5x faster with only a minor degradation in performance.
In particular, \coolname{} uses $40$ and $60$ iterations per frame for tracking \& mapping respectively on Replica, while \coolnamefast{} uses $10$ and $15$ iterations per frame.

\vspace{-1em}

\input{tables/runtime}

\noindent \paragraph{Limitations \& Future Work.}
Although \coolname{} achieves state-of-the-art performance, we find our method to show some sensitivity to motion blur, large depth noise, and aggressive rotation.
We believe a possible solution would be to temporally model these effects and wish to tackle this in future work. 
Furthermore, \coolname{} can be scaled up to large-scale scenes through efficient representations like OpenVDB~\cite{openvdb}. 
Finally, our method requires known camera intrinsics and dense depth as input for performing SLAM, and removing these dependencies is an interesting avenue for the future.

%% file: tables/tracking.tex
\begin{table*}
\centering
\vspace{-1em}
\begin{tabular}{cc}
\setlength{\tabcolsep}{1.0pt}
\vspace{-1em}\input{tables/scannetpp_tracking} & \input{tables/replica_tracking} \\
\input{tables/tum_tracking} &

\input{tables/scannet_tracking}  \\
\end{tabular}
\caption{
\textbf{Online Camera-Pose Estimation Results on Four Datasets} (ATE RMSE \perflower [cm]).
Our method consistently outperforms all the SOTA-dense baselines on ScanNet++, Replica, and TUM-RGBD, while providing competitive performance on Orig-ScanNet~\cite{scannet}.
Best results are highlighted as \colorbox{tabfirst}{\bf first}, \colorbox{tabsecond}{second}, and \colorbox{tabthird}{third}. 
Numbers for the baselines on Orig-ScanNet~\cite{scannet}, TUM-RGBD~\cite{tum-rgbd} and Replica~\cite{replica} are taken from Point-SLAM~\cite{sandstrom2023point}.}
\label{tab:tracking}
\end{table*}

%% file: tables/scannetpp_tracking.tex
\scriptsize
\setlength{\tabcolsep}{5.5pt}

\begin{tabular}[t]{lccc}
\toprule

\textbf{Methods} & \textbf{Avg.} & \texttt{S1} & \texttt{S2} \\

\midrule

Point-SLAM~\cite{sandstrom2023point} & \cellcolor{tabthird}343.8 & \cellcolor{tabthird}296.7 & \cellcolor{tabthird}390.8 \\

ORB-SLAM3~\cite{orbslam3}   &  \cellcolor{tabsecond}158.2 & \cellcolor{tabsecond}156.8 & \cellcolor{tabsecond}159.7 \\

\textbf{\coolname{}}  & \cellcolor{tabfirst}\textbf{1.2} & \cellcolor{tabfirst}\textbf{0.6} & \cellcolor{tabfirst}\textbf{1.9} \\

\bottomrule

\\
\multicolumn{4}{c}{\textbf{ScanNet++~\cite{scannet++}}} \\
\\
\\

\end{tabular}

%% file: tables/replica_tracking.tex
\scriptsize
\setlength{\tabcolsep}{1.5pt}
\begin{tabular}[t]{lcccccccccc}
\toprule

\textbf{Methods} & \textbf{Avg.} & \texttt{R0} & \texttt{R1} & \texttt{R2} & \texttt{Of0} & \texttt{Of1} & \texttt{Of2} & \texttt{Of3} & \texttt{Of4}\\

\midrule

DROID-SLAM~\cite{teed2021droid}      &  \cellcolor{tabsecond}0.38 &  \cellcolor{tabsecond}0.53 &  \cellcolor{tabfirst}\textbf{0.38} &  \cellcolor{tabthird}0.45 &  \cellcolor{tabfirst}\textbf{0.35} &  \cellcolor{tabfirst}\textbf{0.24} &  \cellcolor{tabsecond}0.36 &  \cellcolor{tabsecond}0.33 &  \cellcolor{tabfirst}\textbf{0.43} \\

\cdashmidrule{1-10}

Vox-Fusion~\cite{voxfusion}        &     3.09 &                 1.37 &                      4.70 &                      1.47 &                      8.48 &                      2.04 &                      2.58 &                      1.11 &                      2.94 \\

NICE-SLAM~\cite{nice-slam}         &      1.06 &                0.97 &                      1.31 &                      1.07 &                      0.88 &                      1.00 &                      1.06 &                      1.10 &                      1.13 \\

ESLAM~\cite{eslam}  &                      0.63 &                      0.71 &                      0.70 &                      0.52 &                      0.57 &                      0.55 &                      0.58 &                      0.72 &  \cellcolor{tabthird}0.63 \\

Point-SLAM~\cite{sandstrom2023point} &  \cellcolor{tabthird}0.52 &  \cellcolor{tabthird}0.61 &  \cellcolor{tabthird}0.41 &  \cellcolor{tabsecond}0.37 &  \cellcolor{tabsecond}0.38 &  \cellcolor{tabthird}0.48 &  \cellcolor{tabthird}0.54 &  \cellcolor{tabthird}0.69 &  0.72 \\

\textbf{\coolname{}}                &  \cellcolor{tabfirst}\textbf{0.36} &  \cellcolor{tabfirst}\textbf{0.31} &  \cellcolor{tabsecond}0.40 &  \cellcolor{tabfirst}\textbf{0.29} &  \cellcolor{tabthird}0.47 &  \cellcolor{tabsecond}0.27 &  \cellcolor{tabfirst}\textbf{0.29} &  \cellcolor{tabfirst}\textbf{0.32} &  \cellcolor{tabsecond}0.55 \\

\bottomrule

\\
\multicolumn{10}{c}{\textbf{Replica~\cite{replica}}} \\

\end{tabular}

%% file: tables/tum_tracking.tex
\scriptsize
\setlength{\tabcolsep}{1.5pt}
\begin{tabular}{=l +c +c +c +c +c +c}
\toprule

\multirow{2}{*}{\textbf{Methods}} & \multirow{2}{*}{\textbf{Avg.}} & \texttt{fr1/} &  \texttt{fr1/}  & \texttt{fr1/} & \texttt{fr2/} & \texttt{fr3/}  \\

 &  & \texttt{desk} &  \texttt{desk2} & \texttt{room} & \texttt{xyz}  & \texttt{off.} \\ 

\midrule

Kintinuous~\cite{kintinuous}         &  \cellcolor{tabsecond}4.84 &                      3.70 &                      7.10 &  \cellcolor{tabsecond}7.50 &                      2.90 &  \cellcolor{tabthird}3.00 \\
ElasticFusion~\cite{elasticfusion}   &                      6.91 &  \cellcolor{tabsecond}2.53 &                      6.83 &                      21.49 &  \cellcolor{tabsecond}1.17 &  \cellcolor{tabsecond}2.52 \\
ORB-SLAM2~\cite{orbslam2}            &  \cellcolor{tabfirst}\textbf{1.98} &  \cellcolor{tabfirst}\textbf{1.60} &  \cellcolor{tabfirst}\textbf{2.20} &  \cellcolor{tabfirst}\textbf{4.70} &  \cellcolor{tabfirst}\textbf{0.40} &  \cellcolor{tabfirst}\textbf{1.00} \\

\cdashmidrule{1-7}

NICE-SLAM~\cite{nice-slam}           &                      15.87 &                      4.26 &  \cellcolor{tabthird}4.99 &                      34.49 &                      31.73 &                      3.87 \\
Vox-Fusion~\cite{voxfusion}          &                      11.31 &                      3.52 &                      6.00 &                      19.53 &                      1.49 &                      26.01 \\
Point-SLAM~\cite{sandstrom2023point} &                      8.92 &                      4.34 &  \cellcolor{tabsecond}4.54 &                      30.92 &                      1.31 &                      3.48 \\
\textbf{\coolname{}}                 &  \cellcolor{tabthird}5.48 &  \cellcolor{tabthird}3.35 &                      6.54 &  \cellcolor{tabthird}11.13 &  \cellcolor{tabthird}1.24 &                      5.16 \\

\bottomrule
\\
\multicolumn{7}{c}{\textbf{TUM-RGBD~\cite{tum-rgbd}}} \\
\end{tabular}

%% file: tables/scannet_tracking.tex
\scriptsize
\setlength{\tabcolsep}{1.5pt}
\begin{tabular}{lccccccc}
\toprule

\textbf{Methods} & \textbf{Avg.} & \texttt{0000} & \texttt{0059} & \texttt{0106} & \texttt{0169} & \texttt{0181} & \texttt{0207} \\

\midrule

Vox-Fusion~\cite{voxfusion}        &                      26.90 &                      68.84 &                      24.18 &  \cellcolor{tabsecond}8.41 &                      27.28 &                      23.30 &  \cellcolor{tabthird}9.41 \\

NICE-SLAM~\cite{nice-slam}         &  \cellcolor{tabfirst}\textbf{10.70} &  \cellcolor{tabsecond}12.00 &  \cellcolor{tabthird}14.00 &  \cellcolor{tabfirst}\textbf{7.90} &  \cellcolor{tabfirst}\textbf{10.90} &  \cellcolor{tabsecond}13.40 &  \cellcolor{tabfirst}\textbf{6.20} \\

Point-SLAM~\cite{sandstrom2023point} &  \cellcolor{tabthird}12.19 &  \cellcolor{tabfirst}\textbf{10.24} &  \cellcolor{tabfirst}\textbf{7.81} &  \cellcolor{tabthird}8.65 &  \cellcolor{tabthird}22.16 &  \cellcolor{tabthird}14.77 &                      9.54 \\

\textbf{\coolname{}}                          &  \cellcolor{tabsecond}11.88 &  \cellcolor{tabthird}12.83 &  \cellcolor{tabsecond}10.10 &                      17.72 &  \cellcolor{tabsecond}12.08 &  \cellcolor{tabfirst}\textbf{11.10} &  \cellcolor{tabsecond}7.46 \\

\bottomrule

\\
\multicolumn{8}{c}{\textbf{Orig-ScanNet~\cite{scannet}}} \\

\end{tabular}

%% file: tables/replica.tex
\begin{table}[t]
\centering
\scriptsize
\setlength{\tabcolsep}{1.2pt}

\begin{tabular}{@{}=l +c +c +c +c +c +c +c +c +c +c@{}}
\toprule

\textbf{Methods} & \textbf{Metrics} & \textbf{Avg.} & \texttt{R0} & \texttt{R1} & \texttt{R2} & \texttt{Of0} & \texttt{Of1} & \texttt{Of2} & \texttt{Of3} & \texttt{Of4} \\

\midrule

\multirow{3}{*}{Vox-Fusion~\cite{voxfusion}}
& PSNR \higher        &                      24.41 &  \cellcolor{tabthird}22.39 &                      22.36 &                      23.92 &  \cellcolor{tabthird}27.79 &                      29.83 &  \cellcolor{tabthird}20.33 &  \cellcolor{tabthird}23.47 &  \cellcolor{tabthird}25.21 \\
& SSIM \higher        &                      0.80 &                      0.68 &                      0.75 &  \cellcolor{tabthird}0.80 &  \cellcolor{tabthird}0.86 &                      0.88 &                      0.79 &  \cellcolor{tabthird}0.80 &                      0.85 \\
& LPIPS \perflower        &                      0.24 &  \cellcolor{tabthird}0.30 &  \cellcolor{tabthird}0.27 &                      0.23 &                      0.24 &  \cellcolor{tabthird}0.18 &  \cellcolor{tabthird}0.24 &  \cellcolor{tabthird}0.21 &  \cellcolor{tabthird}0.20 \\

\cdashmidrule{1-11}

\multirow{3}{*}{NICE-SLAM~\cite{nice-slam}} 
& PSNR \higher         &  \cellcolor{tabthird}24.42 &                      22.12 &  \cellcolor{tabthird}22.47 &  \cellcolor{tabthird}24.52 &  \cellcolor{tabsecond}29.07 &  \cellcolor{tabthird}30.34 &                      19.66 &                      22.23 &                      24.94 \\
& SSIM \higher         &  \cellcolor{tabthird}0.81 &  \cellcolor{tabthird}0.69 &  \cellcolor{tabthird}0.76 &  \cellcolor{tabsecond}0.81 &  \cellcolor{tabsecond}0.87 &  \cellcolor{tabthird}0.89 &  \cellcolor{tabthird}0.80 &  \cellcolor{tabthird}0.80 &  \cellcolor{tabthird}0.86 \\
& LPIPS \perflower         &  \cellcolor{tabthird}0.23 &                      0.33 &  \cellcolor{tabthird}0.27 &  \cellcolor{tabthird}0.21 &  \cellcolor{tabthird}0.23 &  \cellcolor{tabthird}0.18 &  \cellcolor{tabthird}0.24 &  \cellcolor{tabthird}0.21 &  \cellcolor{tabthird}0.20 \\

\cdashmidrule{1-11}

\multirow{3}{*}{Point-SLAM~\cite{sandstrom2023point}} 
& PSNR \higher &  \cellcolor{tabfirst}\textbf{35.17} &  \cellcolor{tabsecond}32.40 &  \cellcolor{tabfirst}\textbf{34.08} &  \cellcolor{tabfirst}\textbf{35.50} &  \cellcolor{tabfirst}\textbf{38.26} &  \cellcolor{tabsecond}39.16 &  \cellcolor{tabfirst}\textbf{33.99} &  \cellcolor{tabfirst}\textbf{33.48} &  \cellcolor{tabfirst}\textbf{33.49} \\
& SSIM \higher &  \cellcolor{tabfirst}\textbf{0.98} &  \cellcolor{tabsecond}0.97 &  \cellcolor{tabfirst}\textbf{0.98} &  \cellcolor{tabfirst}\textbf{0.98} &  \cellcolor{tabfirst}\textbf{0.98} &  \cellcolor{tabfirst}\textbf{0.99} &  \cellcolor{tabsecond}0.96 &  \cellcolor{tabfirst}\textbf{0.96} &  \cellcolor{tabfirst}\textbf{0.98} \\
& LPIPS \perflower &  \cellcolor{tabsecond}0.12 &  \cellcolor{tabsecond}0.11 &  \cellcolor{tabsecond}0.12 &  \cellcolor{tabsecond}0.11 &  \cellcolor{tabsecond}0.10 &  \cellcolor{tabsecond}0.12 &  \cellcolor{tabsecond}0.16 &  \cellcolor{tabsecond}0.13 &  \cellcolor{tabfirst}\textbf{0.14} \\

\cdashmidrule{1-11}

\multirow{3}{*}{\textbf{\coolname{}}} 
& PSNR \higher                          &  \cellcolor{tabsecond}34.11 &  \cellcolor{tabfirst}\textbf{32.86} &  \cellcolor{tabsecond}33.89 &  \cellcolor{tabsecond}35.25 &  \cellcolor{tabfirst}\textbf{38.26} &  \cellcolor{tabfirst}\textbf{39.17} &  \cellcolor{tabsecond}31.97 &  \cellcolor{tabsecond}29.70 &  \cellcolor{tabsecond}31.81 \\
& SSIM \higher                          &  \cellcolor{tabsecond}0.97 &  \cellcolor{tabfirst}\textbf{0.98} &  \cellcolor{tabsecond}0.97 &  \cellcolor{tabfirst}\textbf{0.98} &  \cellcolor{tabfirst}\textbf{0.98} &  \cellcolor{tabsecond}0.98 &  \cellcolor{tabfirst}\textbf{0.97} &  \cellcolor{tabsecond}0.95 &  \cellcolor{tabsecond}0.95 \\
& LPIPS \perflower                          &  \cellcolor{tabfirst}\textbf{0.10} &  \cellcolor{tabfirst}\textbf{0.07} &  \cellcolor{tabfirst}\textbf{0.10} &  \cellcolor{tabfirst}\textbf{0.08} &  \cellcolor{tabfirst}\textbf{0.09} &  \cellcolor{tabfirst}\textbf{0.09} &  \cellcolor{tabfirst}\textbf{0.10} &  \cellcolor{tabfirst}\textbf{0.12} &  \cellcolor{tabsecond}0.15 \\

\bottomrule
\end{tabular}
\caption{\textbf{Quantitative Train View Rendering Performance on Replica~\cite{replica}.} \coolname{} is comparable to the SOTA baseline, Point-SLAM~\cite{sandstrom2023point} and consistently outperforms the other dense SLAM methods by a large margin. Numbers for the baselines are taken from Point-SLAM~\cite{sandstrom2023point}. Note that Point-SLAM uses ground-truth depth for rendering.}
\label{tab:replica}
\end{table}

%% file: tables/scannetpp.tex
\begin{table}[t]
\centering

\scriptsize
\setlength{\tabcolsep}{3.5pt}

\begin{tabular}{lcccccccc}
\toprule

\multirow{2}{*}{\textbf{Methods}} & \multirow{2}{*}{\textbf{Metrics}} & \multicolumn{3}{c}{\textbf{Novel View}} & \multicolumn{3}{c}{\textbf{Training View}} \\

 &  & \textbf{Avg.} & \texttt{S1} & \texttt{S2} & \textbf{Avg.} & \texttt{S1} & \texttt{S2} \\

\midrule

\multirow{4}{*}{Point-SLAM~\cite{sandstrom2023point}} & PSNR [dB] \higher & \cellcolor{tabsecond}11.91 & \cellcolor{tabsecond}12.10 & \cellcolor{tabsecond}11.73 & \cellcolor{tabsecond}14.46 & \cellcolor{tabsecond}14.62 & \cellcolor{tabsecond}14.30 \\
& SSIM \higher  & \cellcolor{tabsecond}0.28 & \cellcolor{tabsecond}0.31 & \cellcolor{tabsecond}0.26 & \cellcolor{tabsecond}0.38 & \cellcolor{tabsecond}0.35 & \cellcolor{tabsecond}0.41 \\
& LPIPS \perflower & \cellcolor{tabsecond}0.68 & \cellcolor{tabsecond}0.62 & \cellcolor{tabsecond}0.74  & \cellcolor{tabsecond}0.65 & \cellcolor{tabsecond}0.68 & \cellcolor{tabsecond}0.62 \\
& Depth L1 [cm] \perflower & \cellcolor{tabsecond}\redx & \cellcolor{tabsecond}\redx & \cellcolor{tabsecond}\redx & \cellcolor{tabsecond}\redx & \cellcolor{tabsecond}\redx & \cellcolor{tabsecond}\redx \\

\cdashmidrule{1-8}

\multirow{4}{*}{\textbf{\coolname{}}} & PSNR [dB] \higher & \cellcolor{tabfirst}\textbf{24.41} & \cellcolor{tabfirst}\textbf{23.99} & \cellcolor{tabfirst}\textbf{24.84} & \cellcolor{tabfirst}\textbf{27.98} & \cellcolor{tabfirst}\textbf{27.82} & \cellcolor{tabfirst}\textbf{28.14} \\
& SSIM \higher  & \cellcolor{tabfirst}\textbf{0.88} & \cellcolor{tabfirst}\textbf{0.88} & \cellcolor{tabfirst}\textbf{0.87} & \cellcolor{tabfirst}\textbf{0.94} & \cellcolor{tabfirst}\textbf{0.94} & \cellcolor{tabfirst}\textbf{0.94} \\
& LPIPS \perflower & \cellcolor{tabfirst}\textbf{0.24} & \cellcolor{tabfirst}\textbf{0.21} & \cellcolor{tabfirst}\textbf{0.26} & \cellcolor{tabfirst}\textbf{0.12} & \cellcolor{tabfirst}\textbf{0.12} & \cellcolor{tabfirst}\textbf{0.13} \\
& Depth L1 [cm] \perflower & \cellcolor{tabfirst}\textbf{2.07} & \cellcolor{tabfirst}\textbf{1.91} & \cellcolor{tabfirst}\textbf{2.23} & \cellcolor{tabfirst}\textbf{1.28} & \cellcolor{tabfirst}\textbf{0.93} & \cellcolor{tabfirst}\textbf{1.64} \\

\bottomrule
\end{tabular}
\caption{\textbf{Novel \& Train View Rendering Performance on ScanNet++~\cite{scannet++}.} \coolname{} provides high-fidelity performance on both training views seen during SLAM and held-out novel views from any camera pose. On the other hand, Point-SLAM~\cite{sandstrom2023point} requires ground-truth depth \& performs poorly.}
\label{tab:scannetpp}
\end{table}

%% file: tables/color_depth_ablation.tex
\begin{table}[t]
\centering

\scriptsize
\setlength{\tabcolsep}{5.5pt}

\begin{tabular}{ccccccc}
\toprule

\textbf{Track.} & \textbf{Map.} & \textbf{Track.} & \textbf{Map.} & \textbf{ATE} & \textbf{Dep. L1} & \textbf{PSNR} \\
\textbf{Color} & \textbf{Color} & \textbf{Depth} & \textbf{Depth} & [cm]$\downarrow$ & [cm]$\downarrow$ & [dB]$\uparrow$ \\

\midrule

\redx & \redx & \greencheck & \greencheck & 86.03 & \redx & \redx \\

\greencheck & \greencheck & \redx & \redx & 1.38 & 12.58 & 31.30 \\

\greencheck & \greencheck & \greencheck & \greencheck & \textbf{0.27} & \textbf{0.49} & \textbf{32.81} \\

\bottomrule
\end{tabular}
\caption{\textbf{Color \& Depth Loss Ablation on Replica/Room 0.}}
\label{tab:color_depth_ablation}
\end{table}

%% file: tables/tracking_ablation.tex
\begin{table}[t]
\centering

\scriptsize
\setlength{\tabcolsep}{5.5pt}

\begin{tabular}{ccccccc}
\toprule

\textbf{Velo.} & \textbf{Sil.} & \textbf{Sil.} & \textbf{ATE} & \textbf{Dep. L1} & \textbf{PSNR} \\
\textbf{Prop.} & \textbf{Mask} & \textbf{Thresh.} & [cm]$\downarrow$ & [cm]$\downarrow$ & [dB]$\uparrow$ \\

\midrule

\redx & \greencheck & 0.99 & 2.95 & 2.15 & 25.40 \\

\greencheck & \redx & 0.99 & 115.80 & \redx & \redx \\

\greencheck & \greencheck & 0.5 & 1.30 & 0.74 & 31.36 \\

\greencheck & \greencheck & \textbf{0.99} & \textbf{0.27} & \textbf{0.49} & \textbf{32.81} \\

\bottomrule
\end{tabular}
\caption{\textbf{Camera Tracking Ablations on Replica/Room 0.}}
\label{tab:tracking_ablation}
\end{table}

%% file: figs/rendering.tex
\begin{figure*}[!t]
\centering
{\footnotesize
\setlength{\tabcolsep}{1pt}
\renewcommand{\arraystretch}{0.9}
\newcommand{\sz}{0.12}
\begin{tabular}{ccccccccc}
\rotatebox[origin=c]{90}{PS~\cite{sandstrom2023point}} & 
\raisebox{-0.5\height}{\includegraphics[width=\sz\linewidth]{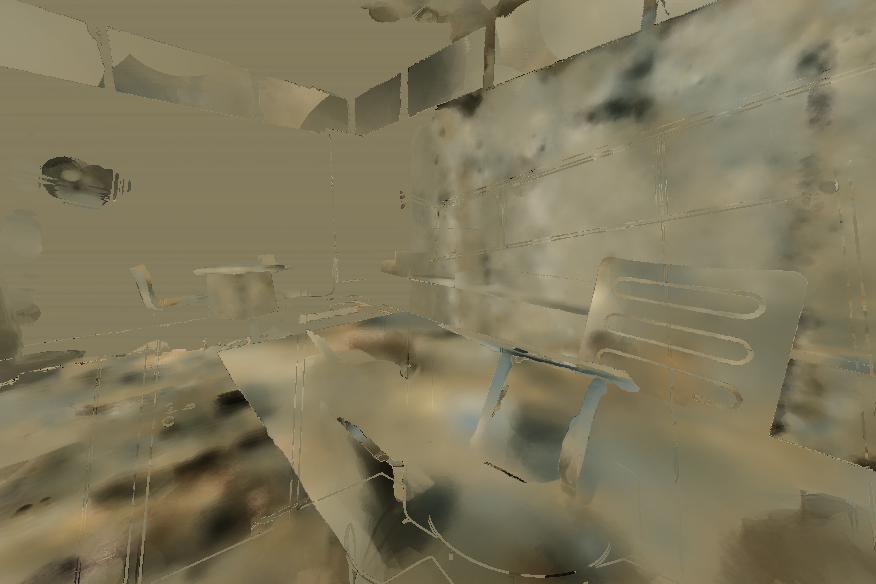}} & 
\raisebox{-0.5\height}{\includegraphics[width=\sz\linewidth]{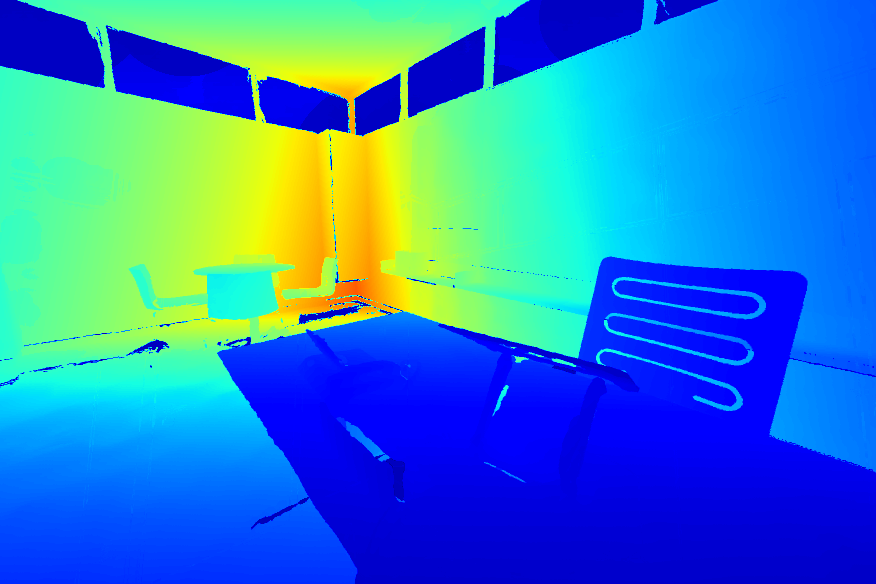}} &
\raisebox{-0.5\height}{\includegraphics[width=\sz\linewidth]{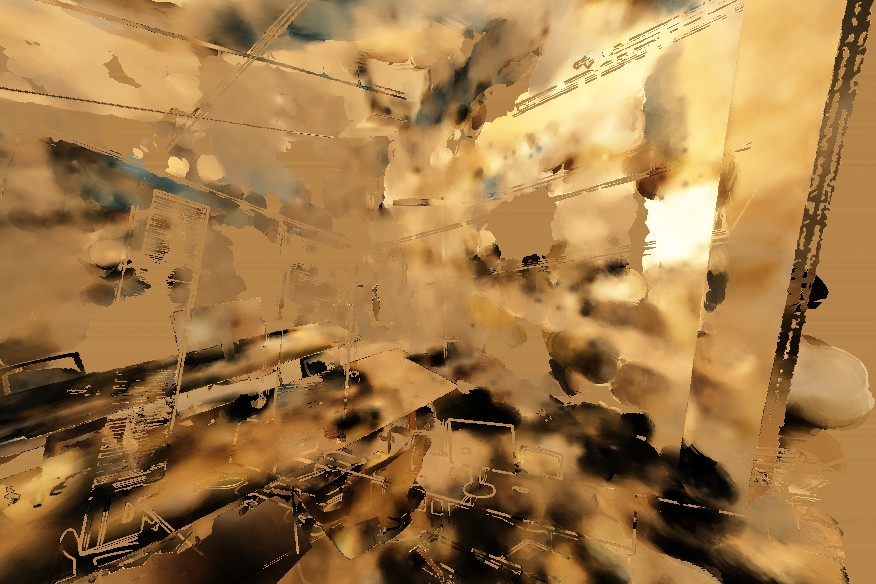}} &
\raisebox{-0.5\height}{\includegraphics[width=\sz\linewidth]{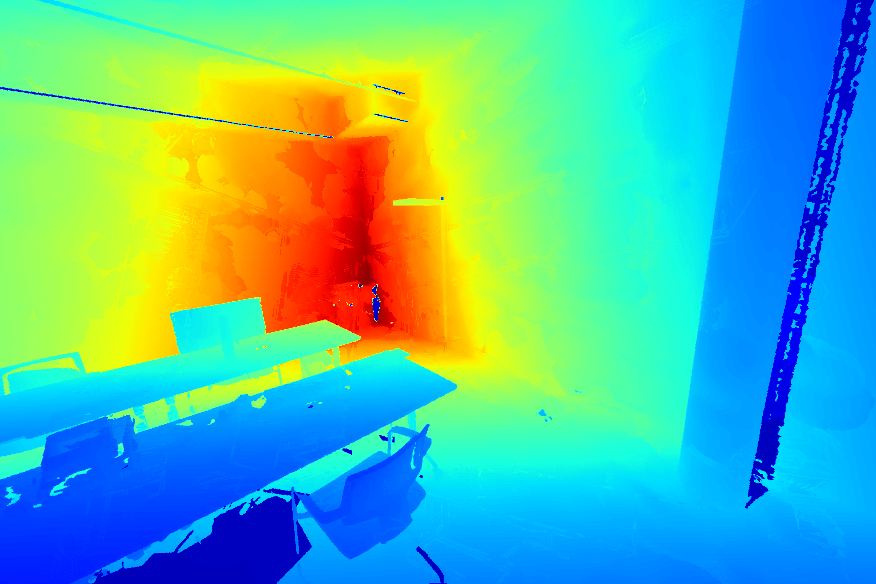}}
& 
\raisebox{-0.5\height}{\includegraphics[width=\sz\linewidth]{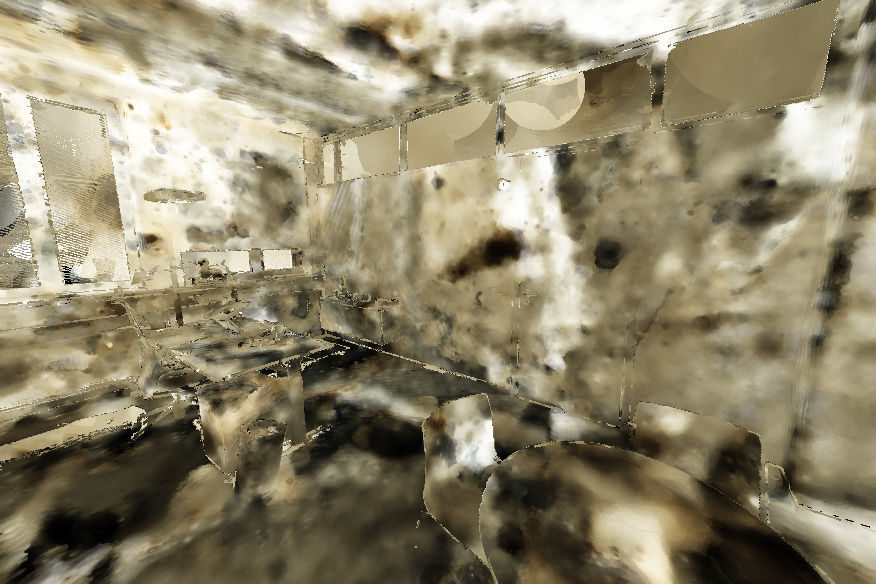}} & 
\raisebox{-0.5\height}{\includegraphics[width=\sz\linewidth]{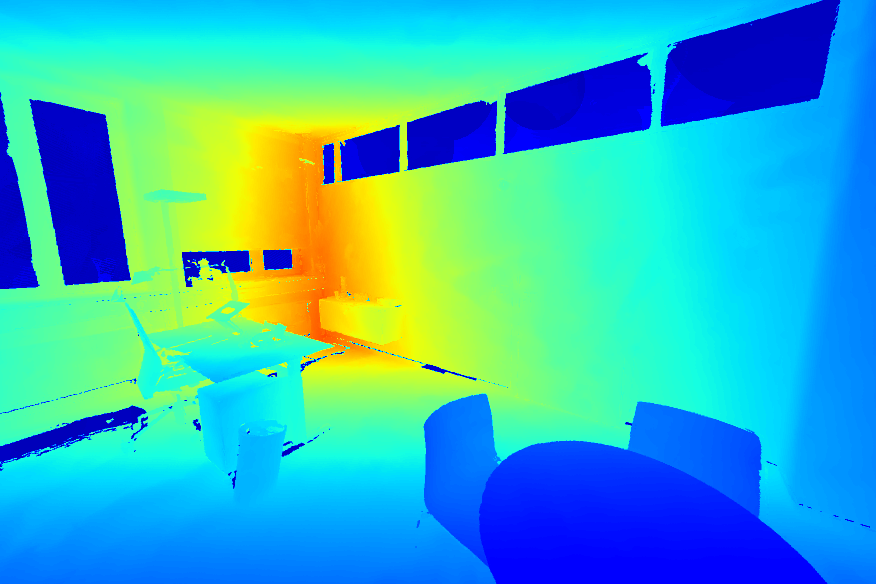}} &
\raisebox{-0.5\height}{\includegraphics[width=\sz\linewidth]{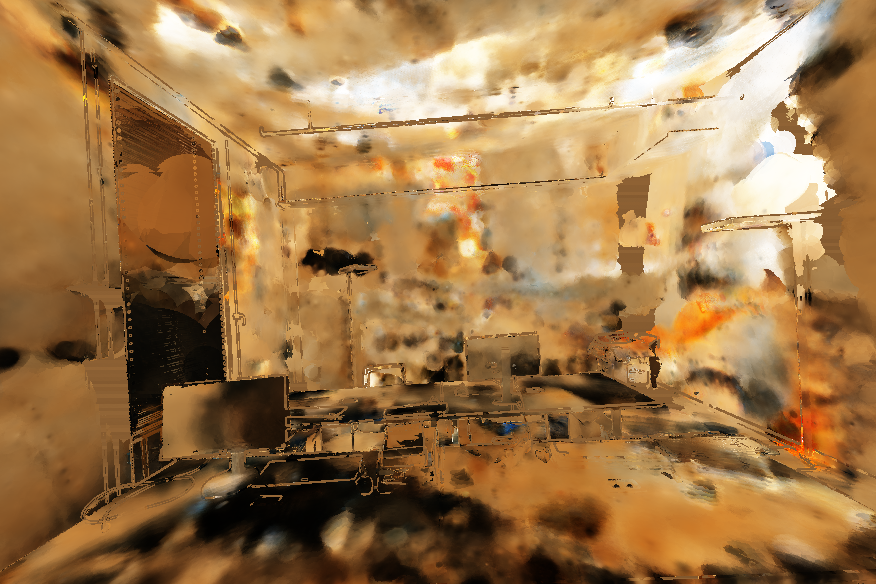}} &
\raisebox{-0.5\height}{\includegraphics[width=\sz\linewidth]{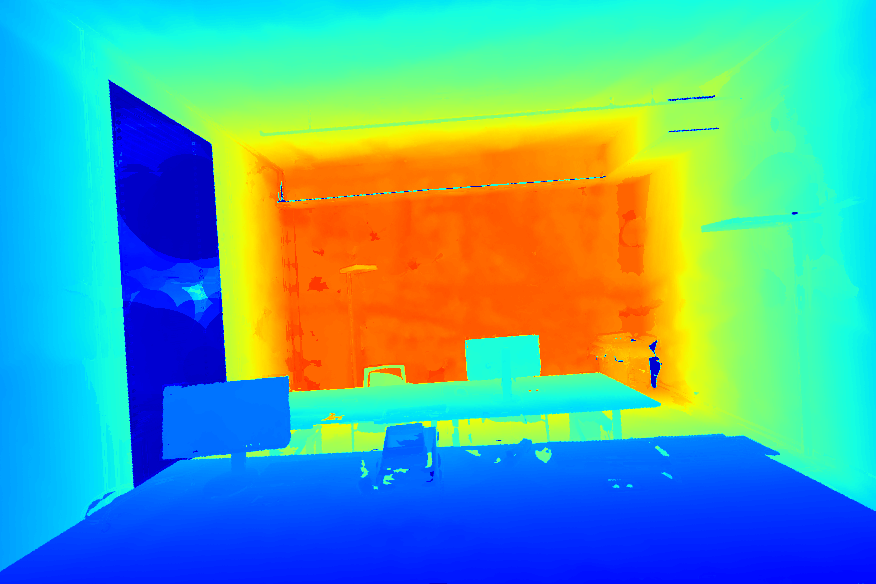}}
\\
\rotatebox[origin=c]{90}{\textbf{Ours}} & 
\raisebox{-0.5\height}{\includegraphics[width=\sz\linewidth]{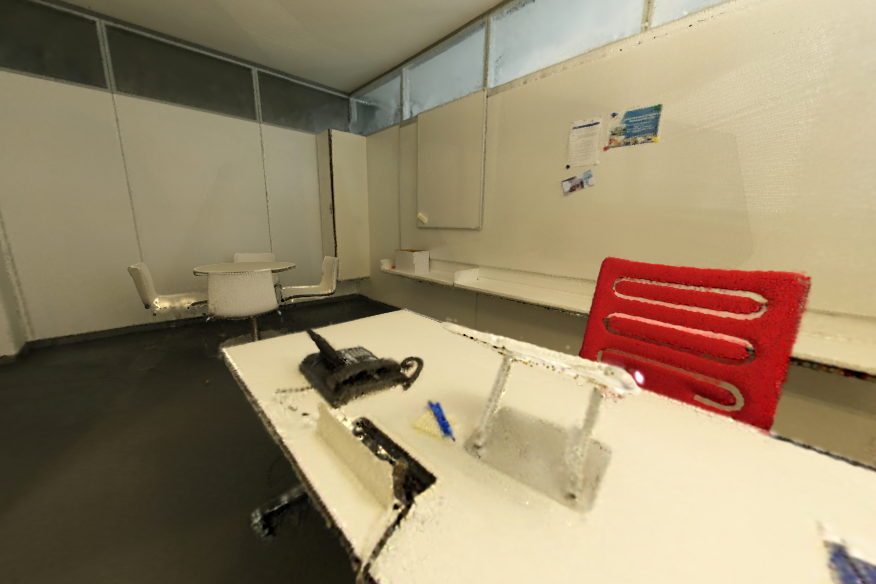}} & 
\raisebox{-0.5\height}{\includegraphics[width=\sz\linewidth]{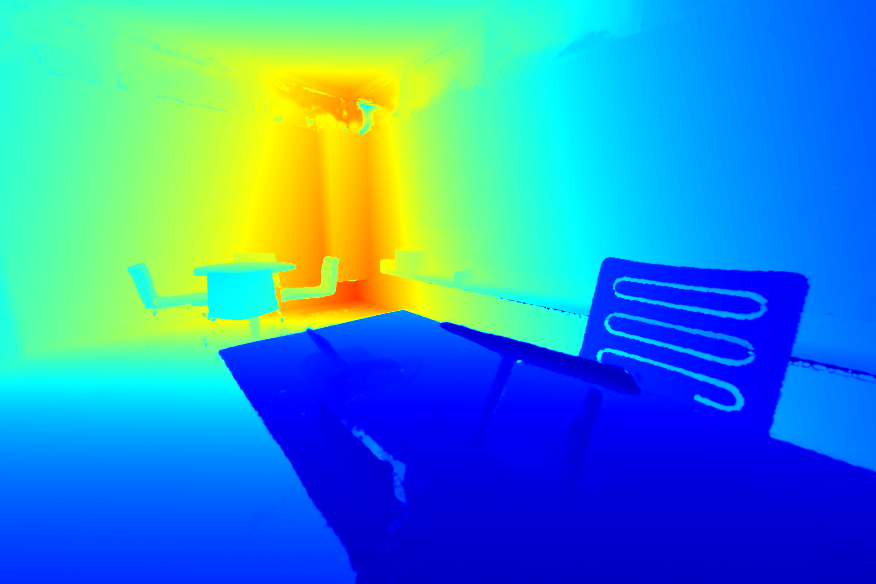}} &
\raisebox{-0.5\height}{\includegraphics[width=\sz\linewidth]{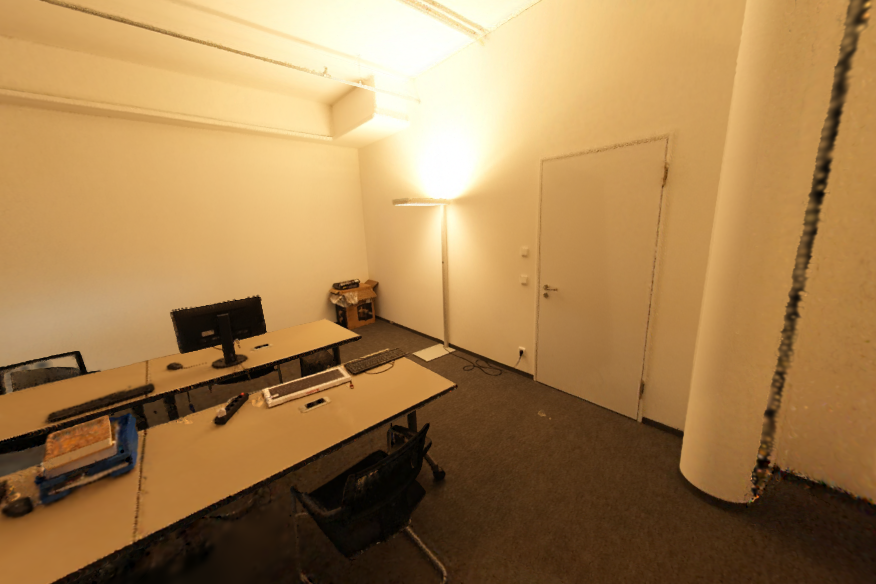}} &
\raisebox{-0.5\height}{\includegraphics[width=\sz\linewidth]{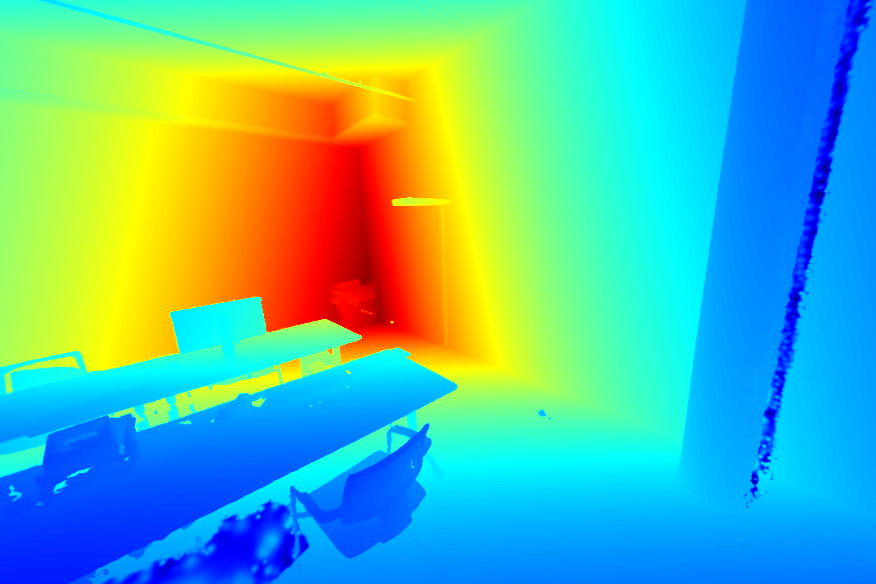}}
& 
\raisebox{-0.5\height}{\includegraphics[width=\sz\linewidth]{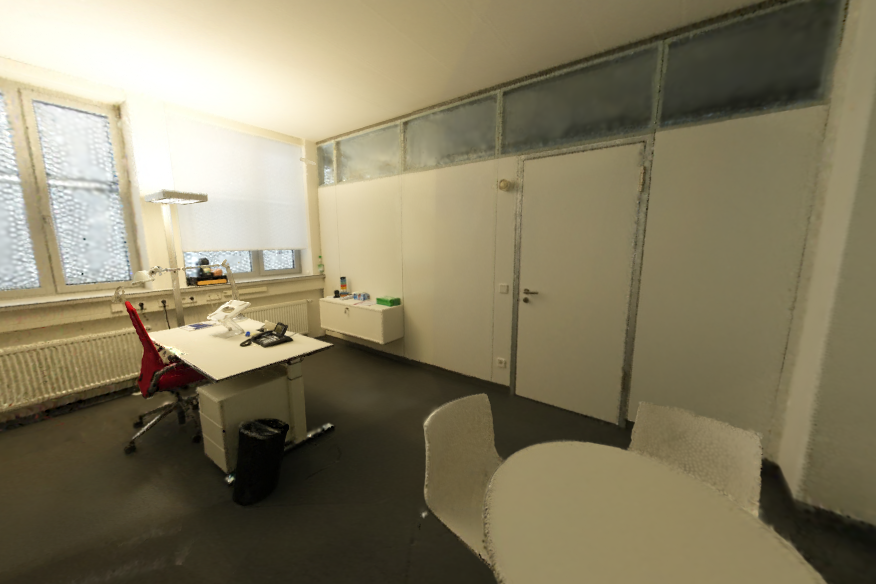}} & 
\raisebox{-0.5\height}{\includegraphics[width=\sz\linewidth]{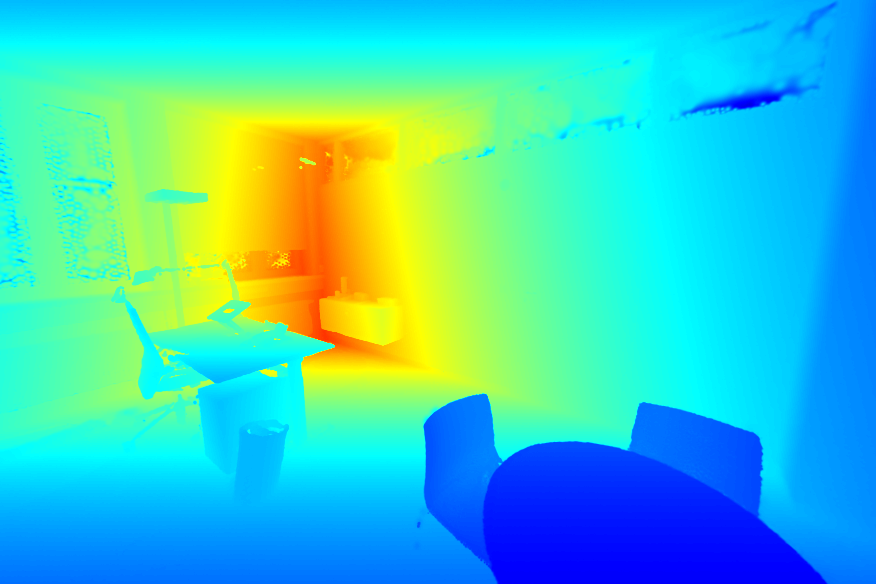}} &
\raisebox{-0.5\height}{\includegraphics[width=\sz\linewidth]{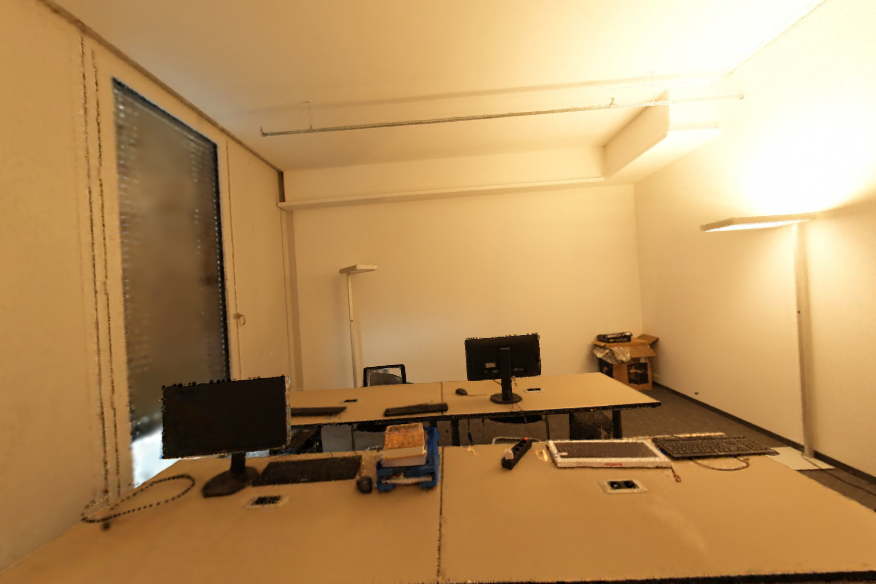}} &
\raisebox{-0.5\height}{\includegraphics[width=\sz\linewidth]{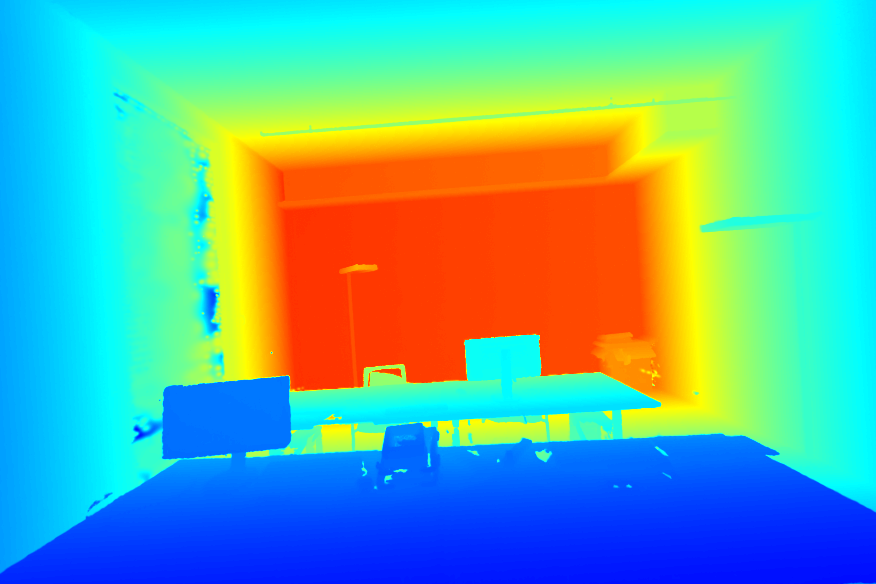}}
\\
\rotatebox[origin=c]{90}{GT} & 
\raisebox{-0.5\height}{\includegraphics[width=\sz\linewidth]{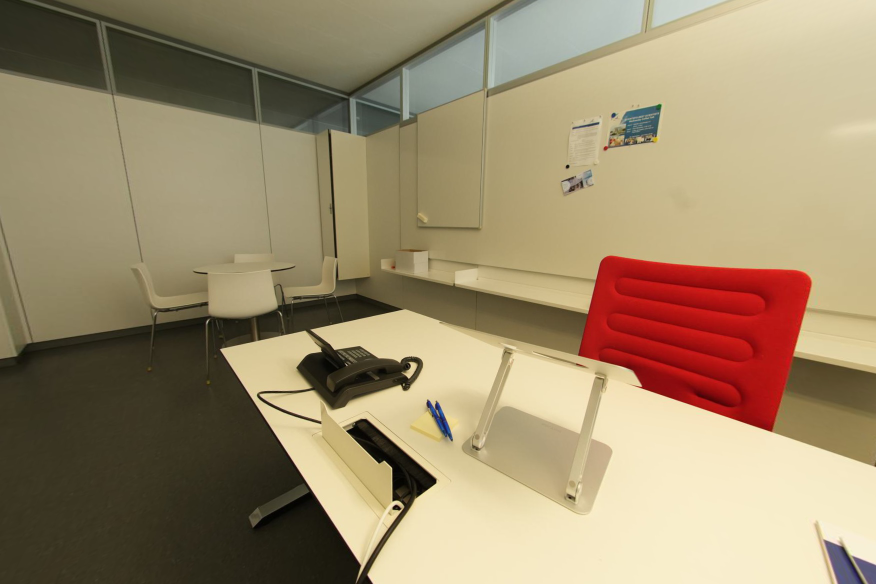}} & 
\raisebox{-0.5\height}{\includegraphics[width=\sz\linewidth]{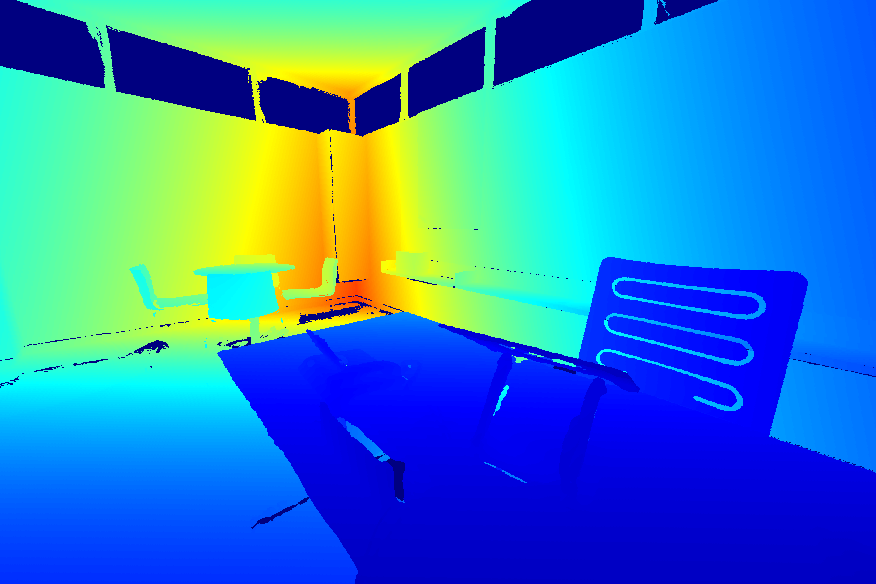}} &
\raisebox{-0.5\height}{\includegraphics[width=\sz\linewidth]{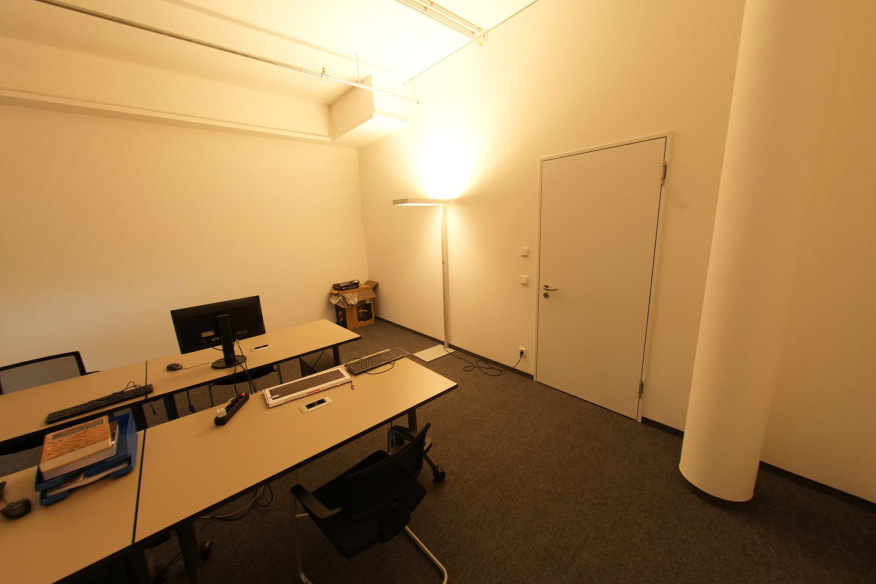}} &
\raisebox{-0.5\height}{\includegraphics[width=\sz\linewidth]{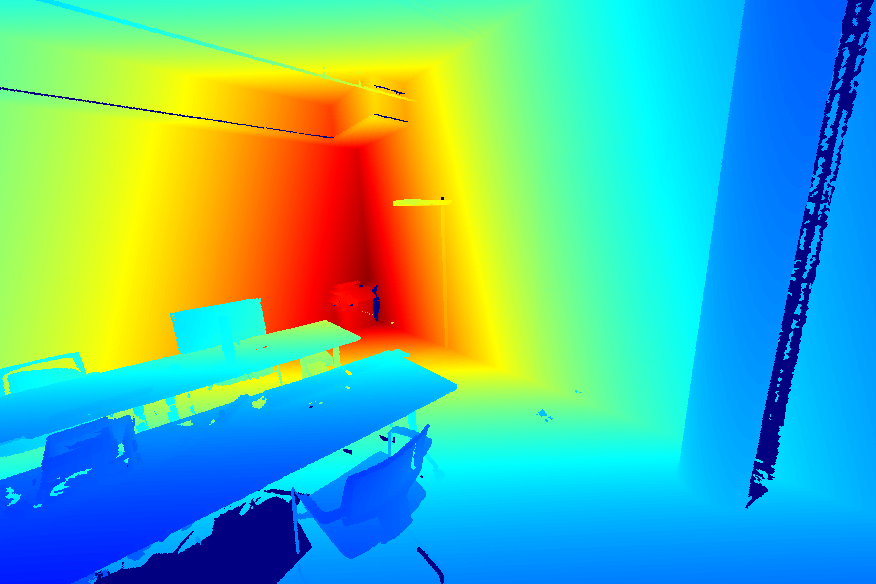}}
& 
\raisebox{-0.5\height}{\includegraphics[width=\sz\linewidth]{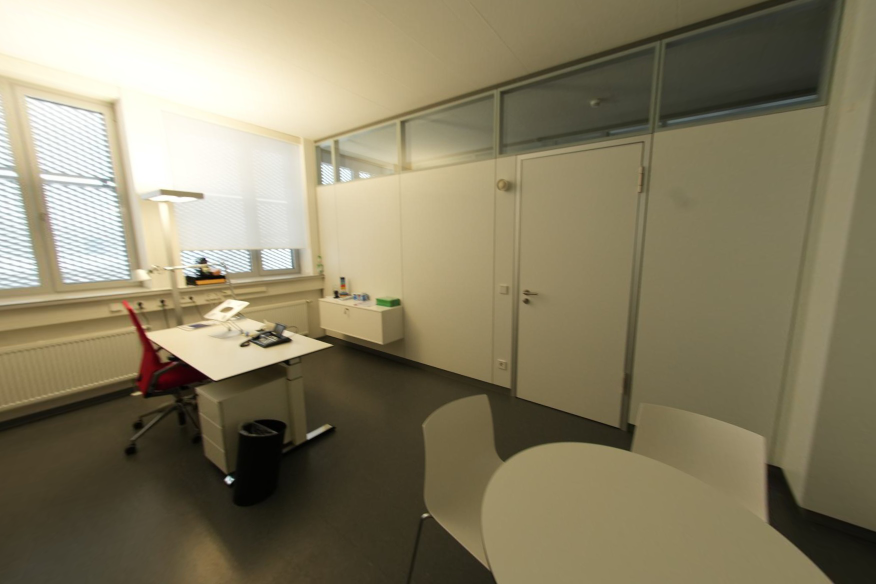}} & 
\raisebox{-0.5\height}{\includegraphics[width=\sz\linewidth]{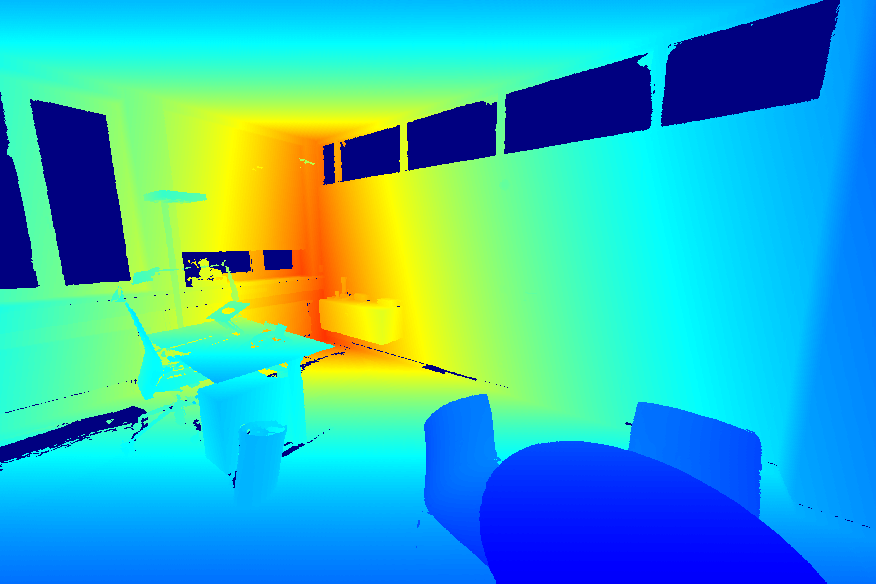}} &
\raisebox{-0.5\height}{\includegraphics[width=\sz\linewidth]{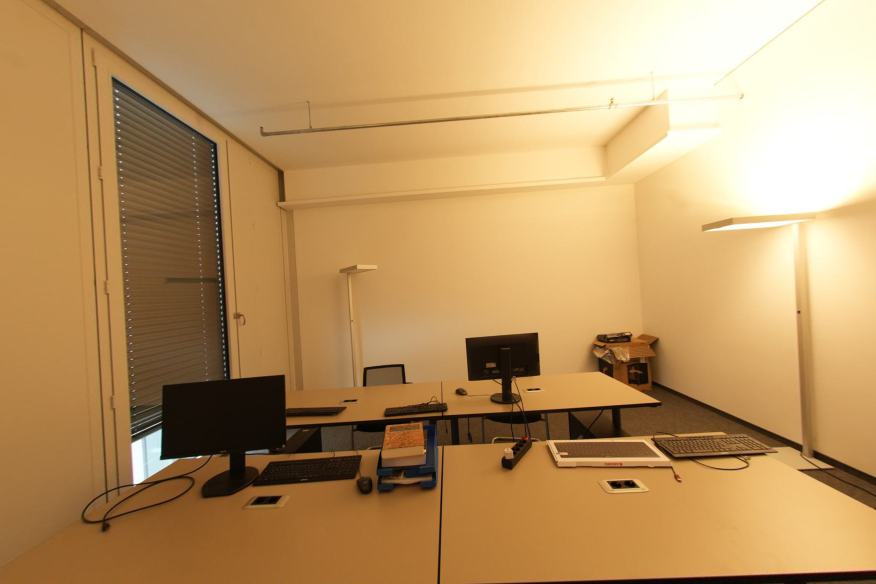}} &
\raisebox{-0.5\height}{\includegraphics[width=\sz\linewidth]{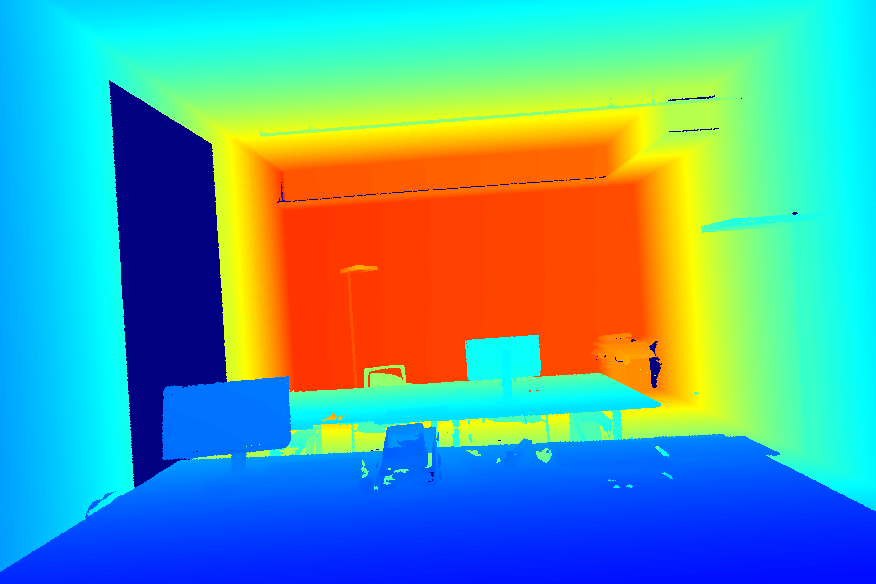}}
\\
& \multicolumn{2}{c}{\texttt{S1}} & \multicolumn{2}{c}{\texttt{S2}} & \multicolumn{2}{c}{\texttt{S1}} & \multicolumn{2}{c}{\texttt{S2}}
\\
& \multicolumn{4}{c}{\texttt{Novel View}} & \multicolumn{4}{c}{\texttt{Train View}} 
\\
\end{tabular}
}
\caption{\textbf{Renderings on ScanNet++~\cite{scannet++}}. Our method, \coolname{}, renders color \& depth for the novel \& train views with fidelity comparable to the ground truth. 
It can also be observed that Point-SLAM~\cite{sandstrom2023point} fails to provide good renderings on both novel \& train view images.
Note that Point-SLAM (PS) uses the ground-truth depth to render the train \& novel view images.
Despite the use of ground-truth depth, the failure of PS can be attributed to the failure of tracking as shown in \cref{tab:tracking}.
}
\label{fig:rendering}
\end{figure*}

%% file: tables/runtime.tex
\begin{table}[!t]
\centering

\scriptsize
\setlength{\tabcolsep}{2.75pt}

\begin{tabular}{lcccccc}
\toprule

\multirow{2}{*}{\textbf{Methods}} & \textbf{Tracking} & \textbf{Mapping} & \textbf{Tracking} & \textbf{Mapping} & \textbf{ATE RMSE} \\
 & \textbf{/Iteration} & \textbf{/Iteration} & \textbf{/Frame} & \textbf{/Frame} & [cm] $\downarrow$ \\

\midrule

NICE-SLAM~\cite{nice-slam}         &  \cellcolor{tabthird}30 ms &                      166 ms &                      1.18 s &  \cellcolor{tabthird}2.04 s & 0.97 \\
Point-SLAM~\cite{sandstrom2023point} &  \cellcolor{tabfirst}\textbf{19 ms} &  \cellcolor{tabthird}30 ms &  \cellcolor{tabsecond}0.76 s &                      4.50 s & \cellcolor{tabthird}0.61 \\
\textbf{\coolname{}}                          &  \cellcolor{tabsecond}25 ms &  \cellcolor{tabsecond}24 ms &  \cellcolor{tabthird}1.00 s &  \cellcolor{tabsecond}1.44 s & \cellcolor{tabfirst}\textbf{0.27} \\
\textbf{\coolnamefast{}}                       &  \cellcolor{tabfirst}\textbf{19 ms} &  \cellcolor{tabfirst}\textbf{22 ms} &  \cellcolor{tabfirst}\textbf{0.19 s} &  \cellcolor{tabfirst}\textbf{0.33 s} & \cellcolor{tabsecond}0.39 \\

\bottomrule
\end{tabular}
\caption{\textbf{Runtime on Replica/R0 using a RTX 3080 Ti.}}
\label{tab:runtime}
\end{table}

%% file: text/05_conclusion.tex
\section{Conclusion}
\label{sec:conclusion}

We present \coolname{}, a novel SLAM system that leverages 3D Gaussians as its underlying map representation to %
enable fast rendering and dense optimization, explicit knowledge of map spatial extent, and streamlined map densification.
We demonstrate its effectiveness in achieving state-of-the-art results in camera pose estimation, scene reconstruction, and novel-view synthesis. 
We believe that \coolname{} not only sets a new benchmark in both the SLAM and novel-view synthesis domains but also opens up exciting avenues, where the integration of 3D Gaussian Splatting with SLAM offers a robust framework for further exploration and innovation in scene understanding. Our work underscores the potential of this integration, paving the way for more sophisticated and efficient SLAM systems in the future.

%% file: supplementary/0_content.tex
\clearpage
\ifcvprfinal
\setcounter{page}{1}
\fi
\maketitlesupplementary

\setcounter{section}{0}
\setcounter{equation}{0}
\setcounter{figure}{0}
\setcounter{table}{0}

\renewcommand{\thesection}{S\arabic{section}}
\renewcommand{\thesubsection}{S\arabic{subsection}}
\renewcommand{\thefigure}{S.\arabic{figure}}
\renewcommand{\thetable}{S.\arabic{table}}
\renewcommand{\theequation}{S.\arabic{equation}}

\section{Overview of Supplementary Material}
\label{sec:overview}

In addition to the qualitative and quantitative results provided in this supplementary PDF, we also provide a public website (\textbf{\href{\webpage}{\color{Green!90}{\webpage}}}) and open-source code (\textbf{\href{\codelink}{\color{Green!90}{\codelink}}}).

The website contains the following material: (i) A 4-minute edited video showcasing our approach \& results, (ii) A thread summarizing the key insights of our paper, (iii) Interactive rendering demos, (iv) Qualitative videos visualizing the SLAM process \& novel-view renderings on ScanNet++~\cite{scannet++}, (v) Qualitative novel view synthesis comparison with Nice-SLAM~\cite{nice-slam} \& Point-SLAM~\cite{sandstrom2023point} on Replica, (vi) Visualizations of the loss during camera tracking optimization, and (vii) Acknowledgment of concurrent work.

Lastly, on our website, we also showcase qualitative videos of online reconstructions using RGB-D data from an iPhone containing a commodity camera \& time of flight sensor.
We highly recommend trying out our online iPhone demo using our open-source code.

\section{Additional Qualitative Visualizations}

\input{figs/scannetpp_recon}

\vspace{-2em}

\paragraph{Visualization of Gaussian Map Reconstruction and Estimated Camera Poses.}
In \cref{fig:recon}, we show visual results of our reconstructed Gaussian Map on the two sequences from ScanNet++. As can be seen, these reconstructions are incredibly high quality both in terms of their geometry and their visual appearance. This is one of the major benefits of using a 3D Gaussian Splatting-based map representation. 

We also show the camera trajectories and camera pose frustums estimated by our method for these two sequences overlaid on the map. One can easily see the large displacement often occurring between sequential camera poses, making this a very difficult SLAM benchmark, and yet one that our approach manages to solve extremely accurately.

\section{Additional Quantitative Results}

\input{tables/rebuttal/distribution_ablation}

\vspace{-1em}

\paragraph{Gaussian Distribution Ablation.}
In \cref{tab:distribution_ablation}, we ablate the benefits of using isotropic (spherical) Gaussians over the anisotropic (ellipsoidal) Gaussians (with no view dependencies/spherical harmonics), as originally used in 3DGS~\cite{3dgs}.
On scenes such as ScanNet++ \texttt{S1}, which contain thin structures, we find a marginal performance difference in SLAM between isotropic and anisotropic Gaussians, where isotropic Gaussians provide faster speed and better memory efficiency.
This supports our design choice of using a 3D Gaussian distribution with fewer parameters.

\input{tables/rebuttal/3dgs_ablation}

\vspace{-1em}

\paragraph{Comparison to 3D Gaussian Splatting.}
To better assess the novel view synthesis performance of \coolname{}, we compare it with the original 3D Gaussian Splatting (3DGS~\cite{3dgs}) using ground-truth poses in \cref{tab:3dgs_ablation}.
We also use \coolname{}'s output for 3DGS \& evaluate rendering performance.
While estimating unknown camera poses online, we can observe that \coolname{} provides novel-view synthesis performance comparable to the original 3DGS~\cite{3dgs} (which requires known poses and performs offline mapping).
This showcases \coolname{}'s potential for simultaneous precise camera tracking and high-fidelity reconstruction.
We further show that the rendering performance of \coolname{} can be slightly improved using 3DGS on the estimated camera poses and map obtained from SLAM.

%% file: figs/scannetpp_recon.tex
\begin{figure}[!h]
    \centering 
    \includegraphics[width=0.99\linewidth]{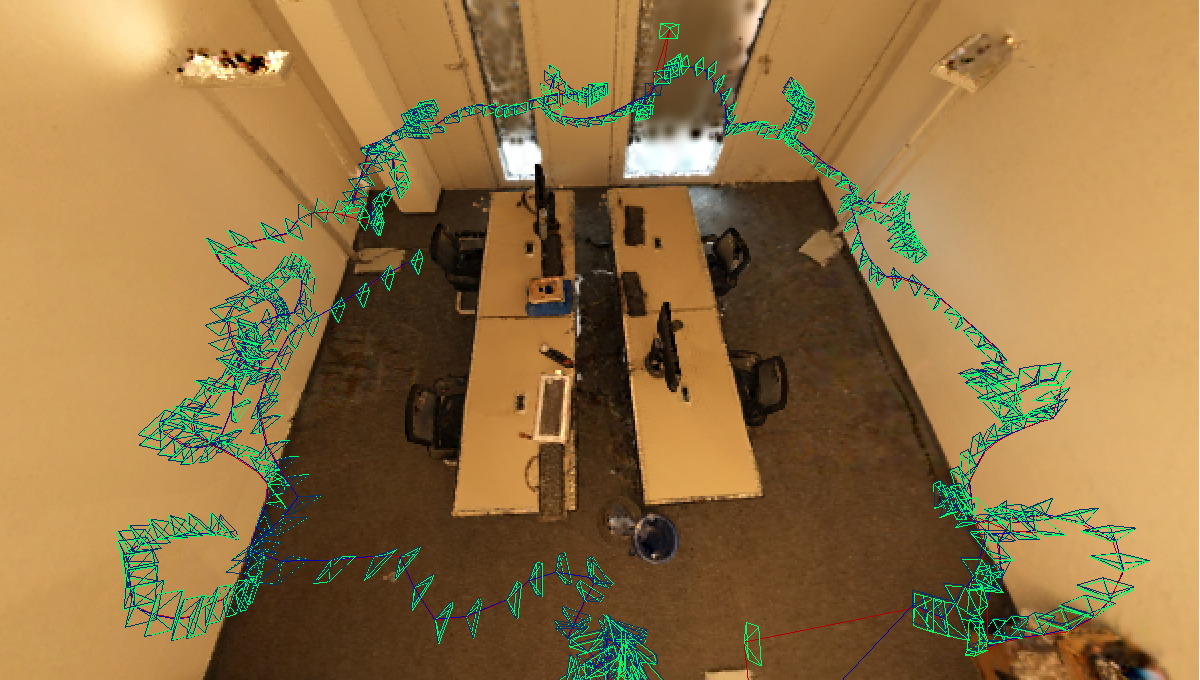}
    \caption{\textbf{Visualization of the reconstruction, estimated and ground truth camera poses on ScanNet++~\cite{scannet++} \texttt{S2}.} 
    It can be observed that the estimated poses from \coolname{} (green frustums \& red trajectory) precisely align with the ground truth (blue frustums \& trajectory) while providing a high-fidelity reconstruction.
    }
    \label{fig:recon}
\end{figure}

%% file: tables/rebuttal/distribution_ablation.tex
\begin{table}[!h]

\centering

\scriptsize
\setlength{\tabcolsep}{5.5pt}

\begin{tabular}{lccccc}
\toprule

 & \textbf{ATE} & \textbf{Training-View} & \textbf{Novel-View} &  \textbf{Time} & \textbf{Memory} \\

\textbf{Distribution} & [cm]$\downarrow$ & \textbf{PSNR} [dB]$\uparrow$ & \textbf{PSNR} [dB]$\uparrow$ & [\%]$\downarrow$ & [\%]$\downarrow$ \\

\midrule

Anisotropic & \textbf{0.55} & \textbf{28.11} & 23.98 & 100 & 100 \\

\textbf{Isotropic} & 0.57 & 27.82 & \textbf{23.99} & \textbf{83.3} & \textbf{57.5} \\

\bottomrule
\end{tabular}

\caption{\textbf{Gaussian Distribution Ablation on ScanNet++ \texttt{S1}.}}
\label{tab:distribution_ablation}

\end{table}

%% file: tables/rebuttal/3dgs_ablation.tex
\begin{table}[!h]

\centering

\scriptsize
\setlength{\tabcolsep}{4pt}

\begin{tabular}{lcccccc}
\toprule

& \multicolumn{3}{c}{\textbf{Novel-View PSNR} [dB]$\uparrow$} & \multicolumn{3}{c}{\textbf{Train-View PSNR} [dB]$\uparrow$} \\

\textbf{Methods} & \textbf{Avg.} & \texttt{S1} & \texttt{S2} & \textbf{Avg.} & \texttt{S1} & \texttt{S2} \\

\midrule

3DGS~\cite{3dgs} (GT-Poses)  & 24.45 & \textbf{26.88} & 22.03 & \textbf{30.78} & \textbf{30.80} & \textbf{30.75} \\

\textbf{Post-\coolname{} 3DGS} & \textbf{25.14} & 25.80 & 24.48 & 27.67 & 27.41 & 27.93  \\

\textbf{\coolname{}} & 24.41  & 23.99 & \textbf{24.84} & 27.98 & 27.82 & 28.14 \\

\bottomrule
\end{tabular}

\caption{\textbf{3D Gaussian Splatting (3DGS) on ScanNet++.}}
\label{tab:3dgs_ablation}

\end{table}